UNSW AUSTRALIA

SCHOOL OF MECHANICAL AND MANUFACTURING ENGINEERING

# Implementation 2D EKF SLAM for Wheeled Mobile Robot

By

Roni Permana Saputra

z5005883

Thesis submitted as a requirement for the degree of Master of Engineering Science Extension in Mechanical Engineering

26th October 2015

Supervisor: Dr. Jose E. Guivant

# Abstract

One of the most important ability in autonomous mobile robot is an ability of the robot to determine its current states (position and heading) in the global map, known as localization. This ability can allow mobile robot to operate autonomously in its workspace without human intervention. Typically, there are two different types of technique in mobile robot navigation. The first one is mobile robot localisation, which relies on the process model of mobile robot and the measurement control input. The second one is mobile robot localisation process, which relies on observation measurement and its knowledge on workspace map. Combination of these techniques provides a higher reliability and a higher accuracy in mobile robot localisation result. Extended Kalman Filter (EKF) is one of the most popular techniques used in the mobile robot localisation process. In some cases, however, mobile robot does not have prior knowledge about the environment. In this case, the mobile robot cannot perform the standard map based localisation operation. Simultaneous Localisation and Mapping (SLAM) has been introduced to tackle this problem. Since then, many research studies have been conducted to improve the performance of SLAM by developing theoretical and conceptual solutions to the SLAM problems. However, in practical implementation, there are still various problems arising due to different implementation and the different set of environments.

The main goal of this project is that the basic EKF-based SLAM operation can be implemented sufficiently for estimating the state of the UGV that is operated in this real environment involving dynamic objects. Several problems in practical implementation of SLAM operation such as processing measurement data, removing bias measurement, extracting landmarks from the measurement data, pre-filtering extracted landmarks and data association in the observed landmarks are observed during the operation of EKF-based SLAM system . In addition, the comparison of EKF-based SLAM operation with dead reckoning operation and Global Positioning System (GPS) are also performed to determine the effectiveness and performance of EKF-based SLAM operation in the real environment.

The systematically processes were performed to achieve these goals. First, the implementation and simulation of the EKF based SLAM operations in different scenario similar with the real scenario were performed. This SLAM operation then was evaluated before the implementation in the real practical implementation. After the validation of SLAM was successfully performed in simulation process, it was then implemented into the real platform with the real data to validate the operation. In this case 'Unmanned Ground Vehicle' (UGV) was used as the platform, which travelling in different scenarios around the main road of University of New South Wales (UNSW) campus. A laser scanner and inertial measurement unit (IMU) sensors, equipped on the UGV provided real measurement data for this process.

The simultaneous and systematic implementation of SLAM operation using UGV as a platform data, which is travelling in different scenarios around the main road of University of New South Wales (UNSW) campus showed that the EKF-based SLAM produced significantly a smaller covariance matrices, a smoother path and a higher accuracy in the estimation of vehicle position in comparison to dead reckoning operation and GPS systems. This data indicated that EKF-based SLAM operation demonstrated a successful operation.

# Certificate of Originality

I, Roni Permana Saputra, hereby declare that this submission is my own work and to the best of my knowledge it contains no materials previously published or written by another person, or substantial proportions of material which have been accepted for the award of any other degree or diploma at UNSW or any other educational institution, except where due acknowledgement is made in the thesis. Any contribution made to the research by others, with whom I have worked at UNSW or elsewhere, is explicitly acknowledged in the thesis.

I also declare that the intellectual content of this thesis is the product of my own work, except to the extent that assistance from others in the project's design and conception in style, presentation and linguistic expression is acknowledged.

Signed    …………………………….

Date      …………………………….

# Acknowledgements

It is a great pleasure for me to have opportunity studying in University of New South Wales. I would like to special thanks to my supervisor **Dr. Jose Guivant**, which is introducing and inspiring me to study about mobile robot localisation and SLAM. I would like to thank also for his advice and constant support throughout this project, so this project become possible.

I would like to express my gratitude to my beloved family, especially my beloved wife for their every time support and encouragement.

# List of Abbreviations

| | |
|---|---|
| 2D | two **D**imensional |
| EKF | **E**xtended **K**alman **F**ilter |
| GPS | **G**lobal **P**ositioning **S**ystem |
| IMU | **I**nertial **M**easurement **U**nit |
| LMS | **L**aser **M**easurement **S**ystem |
| MSE | **M**ean of **S**quared **E**rrors |
| SLAM | **S**imultaneous **L**ocalisation and **M**apping |
| UGV | **U**nmanned **G**round **V**ehicle |

# Nomenclature

X = whole state vector

R = robot state vector

P = covariance matrix

M = map state vector

L = landmarks

# List of Tables

# List of Figures

# Table of Contents

# Chapter

# 1. Introduction

## 1.1. Introduction

Mobile navigation is one of the most interesting topics in robotic research field. Self-localisation of the mobile robot is the ability to localise itself with respect to mapped environment feature in the global workspace to assure autonomous operation. Generally, the robot measures its position and orientation relative to known mapped features and then the robot can determine its expected position in global workspace. In a real world environment, mobile robot navigation have to be able to deal with various problem such as limited access to the maps, strict range limitation, and unpredictable variation that may lead to error and measurement noise during mapping process. Therefore, it is important that the robot can build its own map in order to localize and navigate itself in the sufficiently reliable map of environment. In order to accurately formulate the simultaneous mapping and localisation, the concept of 'Simultaneous Localization and Mapping' (SLAM) has been introduced over the last years. A complete solution to the SLAM problem will allow the mobile robot to operate without human assistance even in an unexplored environment.

The most widely and typical used of SLAM concept is Extended Kalman Filter (EKF) based approach. Basically, in SLAM problem, there are three basic operations, namely states prediction based on robot movement model, states update based on observation, and initialization new observed landmark into the states. While the robot is moving arbitrarily, first, the robot state is predicted by based on movement model. After that, the robot observes its environment to find the possible landmarks. When the robot have prior knowledge about these landmarks states, based on the measurement relative to these particular landmarks, it will update the state estimations and its belief based on this measurement. Meanwhile, if the robot observe new landmark in this process, the robot will estimate these landmark states and add into registered map. EKF estimator play important role in estimating states in both process model and observation based operation. EKF will update the state estimation as well as the covariance and joint covariance of each state component.

## 1.2. Objectives and Significance

There are several works has been done in SLAM concept and implementing this SLAM concept. Many of this implementation is set in static and set environment. The main purpose of this project is to study the basic implementation of EKF-based SLAM in wheeled mobile robot. In this case, 'Unmanned Ground Vehicle' (UGV) is used for obtaining data set from the real environment. The UGV move with different scenario in the main road of The University of New South Wales campus, which containing dynamics objects such as pedestrian and vehicle moving. It is aimed that this basic SLAM operation can be implemented sufficiently for state estimating of the UGV that is operated in this real environment.

## 1.3. Outline

Chapter 2 contains the research background based on literature review on which shows the development and recent research studies on mobile robot localisation and mapping. Several mobile robot navigation technique are describe in this chapter, which also include the advantages and disadvantages of each technique.

Chapter 3 introduce about Kalman Filtering and Extended Kalman Filtering (EKF), which has been widely used for state estimation process in either linear and non-linear process model. This EKF concept will be used as a platform in state estimation process in mobile robot navigation in the next following parts.

Chapter 4 briefly point out the process model that will be used in the state estimation process in this project. These are including the process model of the vehicle movement and the process model of the measurement process. The Jacobian of each process model is also briefly explained, which will be used in linearization process.

Chapter 5 shows the implementation of EKF in mobile robot map-based localisation process in simulation. In this chapter, the result of simulation will be demonstrated to show the performance of EKF-based mobile robot localisation operation.

Chapter 6 explain the whole steps of SLAM operation implemented in simulation process. This part is also developed to validate the whole steps SLAM operation that next will be used in the real practical implementation.

Chapter 7 demonstrate the practical implementation of the EKF-based SLAM operation. In this chapter, the implementation data set obtained from the experiment in the real environments will be used for performing SLAM operation. Several sub-processes that have to be done is also described in this chapter, including accessing measurement data, estimating and removing bias from the measurement data, landmark extraction and pre-filtered landmark. Most importantly, the approach that is used in this project to deal with dynamic objects such as pedestrian is also explained in this chapter. Finally the result of the SLAM operation in this practical implementation is demonstrated with comparison with dead reckoning process and also localisation data obtained from the GPS receiver to validate the SLAM operation that has been implemented.

Finally, chapter 8 concludes the result of the practical implementation of 2D EKF-based SLAM that has been discussed. Last, the possible future studies related to this project will be mentioned in this chapter.

# Chapter

# 2. Background

Localisation is the process of determining the position and orientation in the environment. The localisation is an important ability of any mobile robot navigation, to autonomously and accurately navigates itself in arbitrary environment. Generally, mobile robot navigation processes can be classified into two different basic types, namely relative localisation and absolute localisation method.

## 2.1. Relative Localisation

In this method, mobile robot determines its location based on measurement of its internal dynamic such as its velocity, angular rate, the distance of travel, heading and running time. The most popular technique for relative mobile robot localisation is 'dead reckoning'. This technique can be explained as "the determination, without the aid of external observations, of the position and orientation of a mobile robot from the record of the courses travelled, the distance made, and the known estimated drift" [**1**]. This relative localisation method does not require any external sensors that measure the global states of the robot and complex processing. As a result, this method offers many advantages such as a simple process, relatively inexpensive cost and easy implementation in the real-time process compared to absolute positioning [**2**].

Mobile robot location in this method can be estimated based on recorded dynamics of the mobile robot and its initial recorded location [**3**]. Integral calculation of the change of all dynamics parameter of the mobile robot during the travel is performed to calculate the estimated location. Typically, this relative localisation method process is based on odometry measurement and inertial measurement sensors.

### 2.1.1. Odometry based mobile robot localisation

Acquiring mobile robot position based on odometry sensor is one of the simplest and the most widely used of mobile robot localisation methods [**4**]. Due to its simplicity, this technique can be used for a high-frequency real-time application. This technique also provides high accuracy for mobile robot localisation for a certain amount of travel duration. Moreover, the sensor used in this technique is inexpensive so that it can be implemented for various low-cost applications. In the odometry-based technique, the mobile robot velocity, travelling distance and heading are measured based on information of the number of each wheel revolution of the mobile robot. These revolutions are continuously counted using rotary encoder attached to the mobile robot wheels. Based on this information, velocity and heading rate of this mobile robot is then calculated by integral calculation to estimate the robot location.

### 2.1.2. Inertial-based mobile robot localisation

Another approach for processing relative localisation is by using Inertial Measurement Unit (IMU) sensor. This sensor measures the angular rate of the robot in three local coordinate axes (x, y and z) and also measures the acceleration of this mobile robot in this three coordinate axis [**4**]. This approach has been popular since the development of micro-sensor technology since this sensor is affordable and applicable for various applications. In this approach, the robot localisation can be processed based on an integral calculation of the acceleration to acquire the speed and the gyroscope to measures the angular rate of the robot. Similar to odometry-based approach, the mobile robot position is determined by performing integration combined with rotary encoder based odometry sensor for dead reckoning process on car-like mobile robot [**3**] [**5**]. This method offers a simple and quick process and produces an accurate result for short travel application. Therefore, it is a promising method for real-time application. Moreover, sensors used in this application are relatively inexpensive, so that it is also feasible for mass production application [**5**].

Apart from its various advantages, however, both localisation process based on odometry and based on IMU sensors suffers from both systematic and unsystematic errors. Due to the iterative integration calculation process of the sensors data, the error produced from sensor data will be accumulated and amplified during integral calculation process. Some examples of the main source of error in this technique are the terrain condition and variance of wheels diameter of the vehicle. This system cannot detect and calibrate errors during the process while the robot is travelling. Therefore, this localisation technique is less reliable for a long-term continuous process.

## 2.2. Absolute Localisation

In this method, the robot measures its absolute position and orientation in global coordinate space. This method relies on sensors that provide information about the absolute position of the robot with respect to the global map based on observation process. Typically, these sensors extract information based on observed result of a known object (with known position in global map) around the robot, called landmarks. Therefore, unlike dead reckoning method, this technique is free from drift or systematic error. Theoretically, this technique can be implemented for long-term continuous localisation process. Typical sensors that have been used in this technique include Global Positioning System (GPS), Light Detection and Ranging (LIDAR), Radio Detection and Ranging (RADAR), sonar and visual sensors.

There are two general methods for determining global position for an autonomous vehicle, which are global positioning system and landmark-based positioning system. In the former method, the vehicle determines its position based on global positioning system on satellite constellation. Global Positioning System (GPS) receiver is used to acquire information about the global position of the vehicle. In the latter method, the vehicle measures some known parameters of the environment around it (called feature landmarks). By transforming robot's coordinate position into the global coordination based on measured range and bearing of known landmarks, the mobile robot location can be determined.

### 2.2.1. Global Positioning System (GPS) based localisation

Currently, Global Positioning System (GPS) based localisation has been widely used in vehicle localisation technology. Basically, the GPS system consists of three main parts including satellites, GPS receiver and ground station. . Satellites transmit information of position and current time, ground station controls the satellites and update the information about satellite positions and GPS receiver collects data transmitted by the satellites [**6**]. In vehicle localisation, GPS receiver is used as a sensor that generates information about vehicle location in global map. The vehicle location is determined based on measurement of time to travel of signal from the GPS receiver to several GPS satellites. Theoretically, to determine the two-dimensional position of the vehicle, the measured relative distance from the GPS receiver to three satellites is sufficient. In fact, to get accurate position of the GPS receiver, measured distance from minimum four satellites are required in order to minimise errors in measuring time to travel of the signal [6] [7].

Since it is introduced for public access, applications of GPS based localisation have been widely growth. Localisation process based on GPS provides simple and inexpensive system that can be implemented for mass production application. However, there are several main issues associated with GPS sensor. First, the standard GPS receiver generally has the highest accuracy only at around 3.5 meters that may be not suitable for application requiring high accuracy of the robot position [8]. Secondly, availability of data provided from this sensor is highly dependent on environment [9]. For instance, in urban environment, unreliable data are often produced from GPS due to the signal deterioration and wrong calculation caused by the signal deflection from skyscrapers. Last but not least, GPS receiver does not work in the indoor application.

### 2.2.2. Landmark-based localisation

Landmark-based localisation is another alternative method to determine absolute position of mobile robots. Conceptually, in this method robot's location can be simply determined based on a prior knowledge of the position of landmarks in global map and range and bearing measurement from the robot to the landmarks. However, in practical implementation, this process is not trivial. Many issues have to be overcome to obtain reliable and accurate localisation process, such as identifying good landmarks, data association, effective calculation process and many others. Typical sensors that have been used in landmark-based localisation are including Light Detection and Ranging (LIDAR), Radio Detection and Ranging (RADAR), sonar and visual sensors.

Recently, there has been a growing interest in the use of LIDAR (Light Detection and Ranging) based sensor in robot localisation research studies to enhance the accuracy of the localisation process and dealing with drift [**10**] [**11**] [**12**]. Basically, LIDAR system illuminates target by using laser and then the reflected light is analysed to extract information about the relative distance and bearing of the target object witch respect to the robot. By knowing the location and position of the targeting object on the global map with respect to the target object, the location of the robot on the map can be determined by geometry transformation [**13**]. This sensor is pretty effective for wide range measurement due to good properties in terms of measurement precision and accuracy.

The most popular used of typical LIDAR sensor in this technique, is SICK® laser measurement system (LMS). This sensor is highly accurate with precision value at up to 30 mm [**14**]. Several reports using this sensor have shown the capability of vehicle localisation with precision up to centimetre level [**10**]. For instance, the result of an experiment using LIDAR sensor to simultaneously localisation and mapping of vehicle travelling in urban environment, shows that the maximum error in vehicle localisation on this experiment is only about 35 cm while the vehicle travelled for about 500 m [**10**].

As previously mentioned, in landmark based localisation, the location of vehicle or mobile robot is determined based on prior knowledge of the landmark positions, range and bearing measurement of the landmarks with respect to the vehicle position. Triangulation is one of the most natural methods that can be used to determine the robot global position based on range and bearing measurement of known landmark [**15**]. This method has been implemented for ages in science survey in building map. In order to obtain full measurement of position (x, y) and orientation (θ), three or more measured landmark are required in order to get full states calculation, including position (x, y) and orientation (θ). There are several studies related to mobile robot navigation based on triangulation method, for instance indoor mobile robot navigation based on triangulation of RF signal [**16**]. In this paper, time-to-time RF signal measurement used as the observation measurement for triangulation process to estimate the robot location in the global space coordinate.

Another method that also can be used to determine global position based on range known landmark observation is called trilateration. In this method, position can be calculated based on at least three-range observation of the known landmarks [**17**]. A distinct feature in trilateration method compared to the triangulation method is that bearings of the landmark are not required to calculate position of the vehicle. Nevertheless, both the triangulation and trilateration methods require at least three known landmark measurement for two-dimensional problem and six known landmark measurement for three-dimensional application, which is impractical in some cases. Moreover, there is also a basic shortcoming of using absolute positioning method in robot localization, in which the measurement of observation process is highly depending on the environment characteristics. The changes in the environment feature will really affect performance in the robot localization process [**18**].

## 2.3. Combined Localisation

In recent application, combining both relative positioning and absolute positioning measurement is one of the most popular solution to provide more reliable estimation of robot states by exploiting the advantages of each measurement strength and overcome each sensor limitations [**19**]. Inertial navigation sensor, such as IMU and rotary encoders, combined with global navigation sensor, such as GPS and LIDAR, is one of the most popular integrated data fusion scenarios in vehicle localisation research studies to achieve high integrity and accuracy of localisation system [20].

The challenge now is how to combine information from different sensors measurement, which represent the robot and its environment. Having more than one source of data does not always mean improving the quality of the data but also it can be increasing bias or even destroying the data [21]. It could produce either false negative or false positive result. For instance, when vehicle localisation system detects object in front of the vehicle based on sonar detection but it does not appear in camera detection. In this case, it will result in an error when the system only chooses one sensor and believe in the wrong choice. On the other hand if the system records all possibly detected object by all of the sensors, it would result in two possible outcomes. First, if the detected data is a real object, it means that the system can reduce the blindness. In contrast, if the detected data is a ghost object (error detection), it means we add additional error into the system [21]. In this case, determining proper scenario and methodology in fusing data sensors play an important role in this process. For instance, a complication in fusing data can be minimised by utilising different sensors, which operate a better accuracy at different particular case and region. Therefore, one sensor can cover the limitation case or region of the other sensor and vice versa. The algorithm can manage the use of sensors that operate best in that particular case.

There are several research studies related to the multi-sensor fusion [22]. Multi-sensor fusion methods rely on a probabilistic approach, where the notions of uncertainty and confidence are the common terminology [23] [24]. Kalman filter approach is widely known and widely used for the multi-sensors fusion process. This Kalman filter approach is also popular as the method that used in the mobile robot navigation process. By using Kalman filter approach, it is possible to combine relative positioning navigation and absolute positioning navigation. The extended Kalman filter (EKF) is a modification of Kalman filter algorithm. While Kalman filter can only deal with linear models, the EKF can be used to deal with non-linear models. There are many research studies work successfully on mobile robot navigation based on Kalman filter approach [**18**] [**23**]. Teslic et.al [**25**] presents research work on mobile robot localization operating in a structured environment. In this work, the mobile robot predicts the observation process based on rotary encoder and laser-range-finder. The EKF is used to estimate the states and its covariance in predicted process measurement and updating process measurement. As a result, the mobile robot can achieve satisfactory results in estimating its state and its belief (i.e. covariance matrix).

## 2.4. Simultaneously Localisation and Mapping

In many cases, however, the vehicle cannot rely on landmark-based localization process due to unpredictable variation in the environment. In this case, both environment map and robot's location have to be projected simultaneously. Both estimated robot location and generated map are highly correlated. While the robot is moving, it builds the map simultaneously and uses the generated map to update its states and also update the map [**26**]. To provide the ability for maintaining the mobile robot navigating in the unknown environment, Simultaneous Localization and Mapping (SLAM) concept has been introduced.

In SLAM, to build a reliable map, the vehicle has to have accurate estimation of its states. Meanwhile, to have an accurate estimation of its states, the vehicle needs to have a precise estimation of its surrounding environment.. A better result in SLAM process leads a better ability of autonomous vehicle to navigate in unknown environment autonomously. However, lack of ability in recognizing the environment or consciousness its own states may lead to a very difficult and even almost impossible to operate robot autonomously in unknown environments. These issues are the basic problem in SLAM

There are many research projects on using several approaches to solving problems in SLAM. Among all of various approaches, the Extended Kalman Filter (EKF) based approach and its variances are the most widely used to solve problems in SLAM. Castellanos et.al [**27**] showed a comprehensive experiment of indoor implementation of robot navigation using LabMate$^{TM}$. In this work, a laser rangefinder is used as a sensor for observing the environment around the robot. In another project, Guivant et.al [**28**] presented a successful implementation of EKF-based SLAM algorithm in real time application for an outdoor vehicle.

According to Durrant-Whyte and Bailey, SLAM concept was introduced during the IEEE Robotics and Automation Conference in San Francisco 1986 [**29**]. Since then, many research studies began on finding a fully satisfying solution. Theoretically, problem in SLAM has been solved. However, in reality there are still set of problems arising due to different implementation and the different set of environments. Many researchers have been proposed different approaches to solve different particular problem in practical implementations.

Computational complexity is one of the challenges in solving problem in SLAM, especially for large-scale environment application. This complexity grows remarkably while the new landmark is detected and involved in the SLAM calculation. It increases the size of covariance matrix and states vector. J. Guivant [**30**] investigated the implementation of SLAM process for the large-scale environment [**30**]. The author discusses several ways to obtain efficient SLAM process for real-time large environment application using compressed EKF. Another popular alternative approach to solving computational complexity is by using Fast SLAM method [**31**]. The optimized process by decoupling the robot states estimation and the landmarks state estimation problem. By conditioning the landmark state and robot state estimate, it reduces the computational complexity.

# Chapter

# 3. Kalman Filter

## 3.1. Introduction

Kalman Filter has been widely used in the state of estimation process in both static and dynamics system, which is interfered by noise. The Kalman filter is introduced by Rudolph. E. Kalman [**32**] in 1960 [**33**]. Since then, the Kalman filter has attracted extensive research and application, predominantly in the area of autonomous and navigation. . In Kalman filter, the estimated state on the process is assumed as Gaussian probability density function (PDF), which has mean ($\bar{X}_k$) and covariance (P). In this case, the state is represented as probability of normal distribution. Generally, in the Kalman filter, there are two main operations, namely prediction and update operations. In the state of estimation, firstly the state is predicted based on the dynamic model of the system. Then, the obtained state of prediction is updated or corrected based on measurement process.

In the next following sections, the overview of two types of Kalman filter, namely, discrete version of Kalman filter and extended Kalman filter will be further explained.

## 3.2. Discrete Time Kalman Filter

In the discrete-time linear Kalman filter, the system follows equation 3-1 and 3-2. In this case, the process of model function ($F$) is linear and the uncertainty. $q$ is represented as zero main Gaussian random variable. Estimated state ($X(k+1)$) in this Kalman filter operation, is expressed as PDF. It can be seen from the equation 3-1 that this estimated PDF is resulted from prior PDF $(X(k+1))$, PDF of the uncertainty of the process model ($q(k)$), and the process model itself ($q(k)$). The estimated measurement ($z(k+1)$) is also expressed as PDF, and it is resulted from the measurement model ($HX(k+1)$), and the uncertainty in measurement model ($w(k+1)$), as described in equation 3-2.

$$X(k+1) = FX(k) + q(k) \quad \text{(3-1)}$$

$$z(k+1) = HX(k+1) + w(k+1) \quad \text{(3-2)}$$

Since this Kalman filter process has two main process, prediction and updating process, the following notation are used for expressing the result of state estimations and its covariance.

- $\bar{X}(k+1|k)$ is used to express the state of estimation result in prediction process, based on process model of the system. This notation can be described as the estimated state at time stamp (k+1), based on data obtained in time stamp (k). The notation of the covariance of this estimated state is expressed as $P(k+1|k)$.

- $\bar{X}(k|k)$ is used to express the state of estimated result in updating process based on observation process or measurement. This notation means that the estimated state at time stamp (k) is resulted based on the observation data at the time stamp (k). The notation of the covariance of this estimated state is expressed as $P(k|k)$.

Based on these notations, the prediction step in this Kalman filter step can be performed as in equation 3-3 and 3-4. The $Q(k)$ in the equation 3-4 is expressed in the covariance of noise process model $q(k)$.

$$\bar{X}(k+1|k) = \boldsymbol{F}\bar{X}(k|k) \tag{3-3}$$

$$P(k+1|k) = \boldsymbol{F}P(k|k)\boldsymbol{F}^T + Q(k) \tag{3-4}$$

Another process model considering control input model $(u(k))$ in this system is described in equation 3-5 and 3-6.

$$\bar{X}(k+1|k) = \boldsymbol{F}\bar{X}(k|k) + \boldsymbol{G}u(k) \tag{3-5}$$

$$P(k+1|k) = \boldsymbol{F}P(k|k)\boldsymbol{F}^T + Q(k) \tag{3-6}$$

These equation is also called as 'time update' equations, since this equation predict the state and covariance ahead the time. The next step is updating step based on observation or measurement process. The measurement model of this process can be written as in equation 3-7. In this model, the measurement noise $(w(k+1))$ is also assumed as zero mean Gaussian noise with covariance $(R(k+1))$. Based on this model, the measurement is predicted as in equation 3-8.

$$z(k+1) = \boldsymbol{H}X(k+1) + w(k+1) \tag{3-7}$$

$$\bar{z}(k+1) = \boldsymbol{H}\bar{X}(k+1) \tag{3-8}$$

Based on this measurement model, the updating process is performed by the set of equation as in equation 3-9 to equation 3-13. The equation 3-9 and 3-10 produce the innovation ($v(k+1)$), and the associated covariance ($S$). The kalman gain ($K(k+1)$) then is computed based on this innovation as shown in equation 3-11. Using this Kalman gain ($K(k+1)$) and the innovation ($v(k+1)$), the estimated state and its covariance then can be updated to be $\bar{X}(k+1|k+1)$ and $P(k+1|k+1)$, as describe in equation 3-12 and 3-13, respectively. The expressions in equation 3-12 and 3-13 are then called as 'measurement update' equations.

$$v(k+1) = z(k+1) - \bar{z}(k+1) \tag{3-9}$$

$$S = HP(k+1|k)H^T + R(k+1) \tag{3-10}$$

$$K(k+1) = P(k+1|k)H^T S^{-1} \tag{3-11}$$

$$\bar{X}(k+1|k+1) = \bar{X}(k+1|k) + K(k+1).v(k+1) \tag{3-12}$$

$$P(k+1|k+1) = P(k+1|k) - P(k+1|k).H^T.S^{-1}.H.P(k+1|k) \tag{3-13}$$

## 3.3. Extended Kalman Filter

This discrete version of Kalman filter is only suitable for estimation process of linear system, in reality; however, most of the system is generally nonlinear. Therefore, the Extended Kalman Filter (EKF) approach is introduced to deal with this problem. The model and measurement process of the nonlinear system are written as in equation 3-14 and 3-15.

$$X(k+1) = f(X(k), u(k)) + q(k) \tag{3-14}$$

$$z(k+1) = h(X(k+1)) + w(k+1) \tag{3-15}$$

The sequence of estimation process is similar with the KF process. However, the distinctive process in the EKF is that the linearization is required, to be taken place in every iteration for updating state and covariance. The Jacobian of both system process model and measurement model are used to linearize these systems. These Jacobian matrices are obtained in the current expected value. Thus, the Jacobian of the process model and measurement model can be seen as in equation 3-16 and 3-17. Based on these linearization process, then the state and covariance updates in both prediction and observation step as written in equation 3-18 to equation 3-20. The following step of this EKF process is the same as in the linear KF process.

$$F(k) = \left. \frac{\delta f(X, u)}{\delta X} \right|_{X=\bar{X}(k), u=u(k)} \tag{3-16}$$

$$H(k) = \left. \frac{\delta h(X)}{\delta X} \right|_{X=\bar{X}(k)} \tag{3-17}$$

$$\bar{X}(k+1|k) = f(\bar{X}(k|k), u(k)) \tag{3-18}$$

$$P(k+1|k) = F(k)P(k|k)F(k)^T + Q(k) \tag{3-19}$$

$$z(k+1) = h(\bar{X}(k+1|k)) \tag{3-20}$$

# Chapter 4

# 4. System Models

## 4.1. Robot Kinematics Process Model

The robot kinematics process model constructs the prediction of the robot current states based on the previous states and given control input of the robot. Mathematically, the discrete version of this process model is written as in equation 4-1. The robot states that have to be estimated in this work consist of robot position, (x, y), and robot heading, theta. It can be written as a state vector as it can be seen in the equation 4-2. The control input measuring in this project consists of speed, measured based on odometry sensors, and angular rate, measured based on IMU sensor. These control input can be expressed as in equation 4-3.

$$X_{(k+1)} = f(X_{(k)}, u_{(k)}) \tag{4-1}$$

$$X = \begin{bmatrix} x \\ y \\ \theta \end{bmatrix} \tag{4-2}$$

$$\mathbf{u}_{(k)} = \begin{bmatrix} V_{(k)} \\ \omega_{(k)} \end{bmatrix} \tag{4-3}$$

In discrete time process, the robot state is predicted iteratively in every sample time T. The index k in the equations represent $k^{th}$ sample, which capture in time (k.T).

Since velocity and angular rate control input are measured in this process, the velocity based kinematic model of the robot will be used to predict robot states based on applied control input. The discrete time robot kinematics model for this case is written as in equation 4.

$$X_{(k+1)} = \begin{bmatrix} x_{(k+1)} \\ y_{(k+1)} \\ \theta_{(k+1)} \end{bmatrix} = \begin{bmatrix} x_{(k)} \\ y_{(k)} \\ \theta_{(k)} \end{bmatrix} + T \begin{bmatrix} v_{(k)}.\cos(\theta_{(k)}) \\ v_{(k)}.\sin(\theta_{(k)}) \\ \omega_{(k)} \end{bmatrix} \quad (4\text{-}4)$$

Jacobian of Robot Process Model

$$X_{(k+1)|k} = X_{(k)|k} + T \begin{bmatrix} v_{(k)}\cos(X(3)_{(k)|k}) \\ v_{(k)}\sin(X(3)_{(k)|k}) \\ \omega_{(k)} \end{bmatrix} \quad (4\text{-}5)$$

$$X_{(k+1)|k} = X_{(k)|k} + T \begin{bmatrix} v_{(k)}\cos(X(3)_{(k)|k}) \\ v_{(k)}\sin(X(3)_{(k)|k}) \\ \omega_{(k)} \end{bmatrix} \quad (4\text{-}6)$$

Ideally, by using proper robot kinematics model, the current robot states can be estimated accurately. However, in practical, there are many source of error of this model such as un-modeled friction and noise in sensor measurement. Equation 4-1 can be modified by including noise model as become equation 4-7.

$$X_{(k+1)} = f(X_{(k)}, u_{m(k)} + \delta u_{(k)}) \equiv f(X_{(k)}, u_{m(k)}) + \delta f \quad (4\text{-}7)$$

In this case, $\delta f$ is represent as noise model, which is approximated as Gaussian noise with zero mean. The source of this noise is from the measurement process, so the covariance of this system noise can be calculated based control measurement covariance as it can be seen in equation 4-8.

$$Q_f = F_u . Q_u . F_u^T \qquad (4\text{-}8)$$

In which:

$Q_f$ = Process noise covariance

$Q_u$ = Control input measurement covariance

$F_u = \dfrac{\delta f}{\delta(v,\omega)}$

The Jacobian matrices of process model with respect to the robot states $F_{X_v}$ and control input measurement $F_u$ can be seen in equation 4-9 and 4-10, respectively.

$$F_u = \frac{\delta f}{\delta(x,y,\theta)} = \begin{bmatrix} 1 & 0 & -T.\text{v}_{(k)}.\sin(\theta_{(k)}) \\ 0 & 1 & T.\text{v}_{(k)}.\cos(\theta_{(k)}) \\ 0 & 0 & 1 \end{bmatrix} \qquad (4\text{-}9)$$

$$F_u = \frac{\delta f}{\delta(v,\omega)} = \begin{bmatrix} T.\cos(X(3)_{(k)}) & 0 \\ T.\sin(X(3)_{(k)}) & 0 \\ 0 & T \end{bmatrix} \qquad (4\text{-}10)$$

## 4.2. Observation Measurement Model

Observation measurement model represents the expected measurement value based on mathematical model of the measurement components, and predicted robot states when the robot measures the landmarks. Mathematically, this observation measurement model can be written as in equation 4-11. In this project, the robot equipped by laser scanner sensors that produce range and bearing measurement of the landmarks with respect to the robot. Thus, these measurement components can be expressed as measurement vector as in equation 4-12. Equation 4-13 shows the mathematical expression to calculate the expected range and bearing measurement based on predicted robot states and landmark location.

$$y_i = h(X_{v,(k)}, L_{i,(k)}) \tag{4-11}$$

$$h(X_{v,(k)}, L_{i,(k)}) = \begin{bmatrix} range(r) \\ bearing(\alpha) \end{bmatrix} \tag{4-12}$$

$$h(X_{v,(k)}, L_{i,(k)}) = \begin{bmatrix} \sqrt{(x_l(k) - x(k))^2 + (y_l(k) - y(k))^2} \\ tan^{-1}\left(\frac{y_l(k) - y(k)}{x_l(k) - x(k)}\right) + \theta(k) + \pi/2 \end{bmatrix} \tag{4-13}$$

The Jacobian matrices of this observation measurement model with respect to the robot state $H_{X_v}$ and observed landmark states $H_{L_i}$ can be seen in equation 4-15 and 4-17, respectively.

$$H_{X_v} = \frac{\partial h(X_{v,(k)}, L_{i,(k)})}{\partial X_v} = \begin{bmatrix} \frac{\partial h1}{\partial x} & \frac{\partial h1}{\partial y} & \frac{\partial h1}{\partial \theta} \\ \frac{\partial h2}{\partial x} & \frac{\partial h2}{\partial y} & \frac{\partial h2}{\partial \theta} \end{bmatrix} \tag{4-14}$$

$$H_{X_v} = \begin{bmatrix} \frac{x_l - x(k)}{\sqrt{(x_l - x(k))^2 + (y_l - y(k))^2}} & \frac{y_l - y(k)}{\sqrt{(x_l - x(k))^2 + (y_l - y(k))^2}} & 0 \\ \frac{y_l - y(k)}{(x_l - x(k))^2 + (y_l - y(k))^2} & \frac{x_l - x(k)}{(x_l - x(k))^2 + (y_l - y(k))^2} & 1 \end{bmatrix} \tag{4-15}$$

$$H_{L_i} = \frac{\partial h(X_{v,(k)}, L_{i,(k)})}{\partial L_i} = \begin{bmatrix} \frac{\partial h1}{\partial x_l} & \frac{\partial h1}{\partial y_l} \\ \frac{\partial h2}{\partial x_l} & \frac{\partial h2}{\partial y_l} \end{bmatrix} \tag{4-16}$$

$$H_{L_i} = \begin{bmatrix} \dfrac{x(k) - x_l}{\sqrt{(x_l - x(k))^2 + (y_l - y(k))^2}} & \dfrac{y(k) - y_l}{\sqrt{(x_l - x(k))^2 + (y_l - y(k))^2}} \\ \dfrac{y(k) - y_l}{(x_l - x(k))^2 + (y_l - y(k))^2} & \dfrac{x(k) - x_l}{(x_l - x(k))^2 + (y_l - y(k))^2} \end{bmatrix} \quad (4\text{-}17)$$

## 4.3. Inverse Observation Model

One distinct feature in SLAM operation compare to regular EKF-based localisation is landmark initialisation. New landmark states are estimated based on observation measurement results and prior robot states. The mathematical model to estimate new landmarks states is known as 'Inverse Observation Model'. In equation 4-18, it can be seen the function of inverse observation model.

$$X_{L_{n+1}} = g(X_{v,(k)}, z_{(k)}) \quad (4\text{-}18)$$

$$g(X_{v,(k)}, z_{(k)}) = \begin{bmatrix} x_v + r.\cos(\alpha_k + \theta_{v,(k)}) \\ y_v + r.\sin(\alpha_k + \theta_{v,(k)}) \end{bmatrix} \quad (4\text{-}19)$$

Equation 4-21 and equation 4-23 show the Jacobian matrices of the inverse observation model function, with respect to robot states $X_v$ and observation measurement z, respectively.

$$G_{X_v} = \dfrac{\partial g(X_{v,(k)}, z_{(k)})}{\partial X_v} = \begin{bmatrix} \dfrac{\partial g1}{\partial x} & \dfrac{\partial g1}{\partial y} & \dfrac{\partial g1}{\partial \theta} \\ \dfrac{\partial g2}{\partial x} & \dfrac{\partial g2}{\partial y} & \dfrac{\partial g2}{\partial \theta} \end{bmatrix} \quad (4\text{-}20)$$

$$G_{X_v} = \begin{bmatrix} 1 & 0 & -r.\sin(\alpha_k + \theta_{v,(k)}) \\ 0 & 1 & r.\cos(\alpha_k + \theta_{v,(k)}) \end{bmatrix} \quad (4\text{-}21)$$

$$G_z = \frac{\partial g(X_{v,(k)}, Z_{(k)})}{\partial z(r, \alpha)} = \begin{bmatrix} \frac{\partial g1}{\partial r} & \frac{\partial g1}{\partial \alpha} \\ \frac{\partial g2}{\partial r} & \frac{\partial g2}{\partial \alpha} \end{bmatrix}$$  (4-22)

$$H_{L_i} = \begin{bmatrix} \cos(\alpha_k + \theta_{v,(k)}) & -r.\sin(\alpha_k + \theta_{v,(k)}) \\ \sin(\alpha_k + \theta_{v,(k)}) & r.\cos(\alpha_k + \theta_{v,(k)}) \end{bmatrix}$$  (4-23)

# Chapter

# 5. 2D EKF-based Localisation

Basic cycle of EKF localisation can be seen in figure 5-1. In general, this cycle consists of two main steps, prediction step based on process model and update step based on map and observation. Point landmarks based approach is used according to the range and bearing measurement of landmarks relative to the robot.

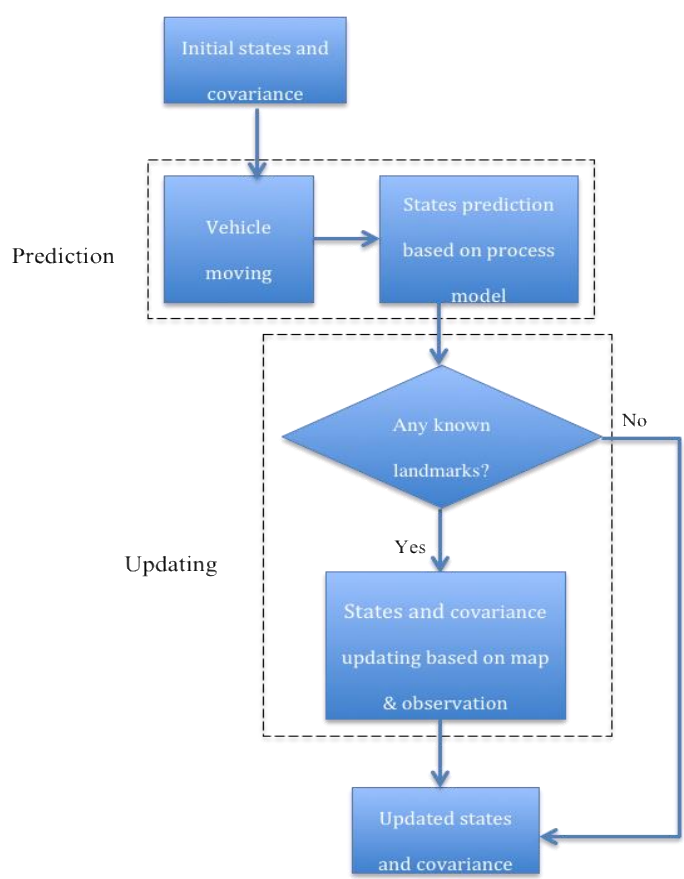

**Figure 5-1**. General operation step for implementation 2D SLAM

In this localisation problem, the estimated states are only robot states, which consist of two dimensional position (x and y), and robot heading (θ). Thus, the estimated states for this problem can be written as a static state vector as in equation 5-1.

$$\bar{X} = \bar{X}_v = \begin{bmatrix} x \\ y \\ \theta \end{bmatrix} \quad (5\text{-}1)$$

## 5.1 General Operation for 2D Landmark based Localisation

### 5.1.1. Step 0. Robot's States Initialization

At the initial time, the robot's states are assumed to be known exactly in a certain position and heading. In this case, for instance the initial position of the robot is assumed in (0,0) and the initial heading is 0.5π. The initial state vector of the robot then can be written as in equation 5-2. Since the robot states are assumed to be known exactly, the initial covariance matrix of this robot state vector is also assumed as zero, as can be seen in equation 5-3.

$$\bar{X}_0 = \bar{X}_{v,0} = \begin{bmatrix} 0 \\ 0 \\ \frac{\pi}{2} \end{bmatrix} \quad (5\text{-}2)$$

$$P_0 = \begin{bmatrix} 0 & 0 & 0 \\ 0 & 0 & 0 \\ 0 & 0 & 0 \end{bmatrix} \quad (5\text{-}3)$$

### 5.1.2. Step 1. Prediction Step

*1. States prediction*

The state prediction process is performed when the robot is moving. Current robot states ($X_{v,(k+1)}$) are predicted based on the prior robot states ($X_{v,(k)}$), and measured applied control input of the robot (u). Since the measured control input is polluted by noise, noise model (N) is also included into this state prediction operation. Thus, the new estimated of robot state ($X_v$) is predicted as in equation (5-4).

$$\overline{X}_v \leftarrow f_{X_v}(\overline{X}_{v,(k)}, u, \overline{N}) \tag{5-4}$$

Equation (5-4) corresponds to the robot process model equation (4-4) of robot based on its previous state ($X_v$) control input (u), and noise model (N). In this process, the noise is modelled as white noise, so that the $\overline{N}$ is equal to zero.

*2. Updating robot covariance*

The robot state covariance is updated based on the equation of Jacobian process model with respect to the robot state vector ($F_{X_v}$) at current time iteration. The noise model covariance ($Q_u$) is also updated based on the Jacobian of robot process model relative to the control input vector ($F_u$). Equation 5-5 shows the covariance matrix updating process in this step. The Jacobian matrices ($F_{X_v}$ and $F_u$) in this updating covariance process correspond to the equation 4-8 and equation 4-10, respectively.

$$P \leftarrow F_{X_v} P F_{X_v}^T + F_u . Q_u . F_u^T \tag{5-5}$$

### 5.1.3. Step 2. Updating Process Based on Landmark Observation

While the robot is moving, the robot observes landmarks around the robot using laser scanner sensor. The sensor measures range and bearing of detected point landmarks in every cycle. When known landmarks are detected, the estimated state vector of the robot is then updated based on difference in actual measurement ranges and bearings of detected landmarks, and expected measurement of ranges and bearings of correspond landmarks based on measurement model equation. Equation 5-6 shows the error calculation between real measurement value and expected measurement value. In this equation, $h_i(X_v, L_i)$ corresponds to the observation measurement model equation 4-13 in chapter 4.

$$\bar{z} = y_i - h_i(X_v, L_i) \tag{5-6}$$

Based on this error measurement model, the EKF-based states updating process can be determined based on set of equation 5-7 into equation 5-10, in which K is Kalman gain of this state updating process. The covariance matrix of the state vector is also updated based on Kalman gain value (K) (see equation 5-11). In this set of equations, $H_{X_v}$ and R correspond to the Jacobian matrix of observation measurement model with respect to state vector 4-15 and noise observation measurement model, respectively.

$$\bar{z} = y_i - h_i(X_v, L_i) \tag{5-7}$$

$$S = H_{X_v} P H_{X_v}^T + R \tag{5-8}$$

$$K = P H_{X_v}^T S^{-1} \tag{5-9}$$

$$\overline{X_V} \leftarrow \overline{X_V} + K\overline{z} \tag{5-10}$$

$$P \leftarrow P - KSK^T \tag{5-11}$$

## 5.2 Simulation

Figure 1 shows the simulation framework in MATLAB that has been developed to simulate and examine several scenarios of this EKF-based robot localisation operation. Using this simulation framework, several scenarios can be simulated to evaluate the performance of EKF-localisation operation, including various noises in measurement control input and laser scanner sensors, various input control for the robot, various landmark and path scenario.

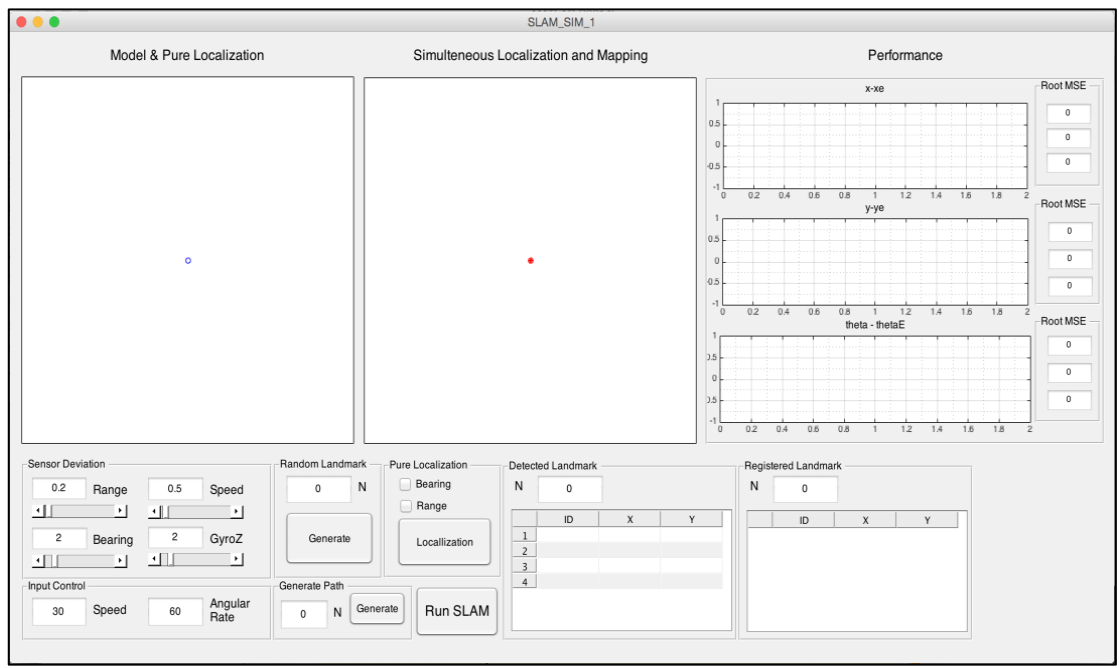

**Figure 5-2.** Designed simulation framework to simulate and test the operations.

Figures 5-3 to 5-6 show the simulation result of four different state estimation processes, including dead reckoning and EKF localization with full and partial observation. The list of parameter setting for this simulation process can be seen in the Table 1. According to these figures, it can be briefly seen that the path estimation as a result of state estimation in EKF localization is more reliable and more accurate in comparison to the dead reckoning localisation in the case of high amount of noise in control input measurement and long-term travelling operation. In EKF-based localisation simulation process as it can be seen in figures 5-4 to 5-6, it is simulated that the robot can maintain the estimation of its state as close as the real value given on the simulation. In contrast, the path estimation obtained by dead reckoning technique is drifted while the robot is moving, as it can be seen in figure 5-3. Therefore, these can be used as rapid indicator that the EKF-based robot localisation in this simulation is performing well.

**Table 5-1.** Parameter setting for the simulation EKF-based localisation process

| Parameter | Value |
| --- | --- |
| Velocity StDev | 0.5 m/s |
| Angular rate StDev | 2 degree/s |
| Range measurement StDev | 0.2 m |
| Bearing measurement StDev | 2 degree |
| Maximum input speed | 30 m/s |
| Maximum input angular rate | 60 degree/s |

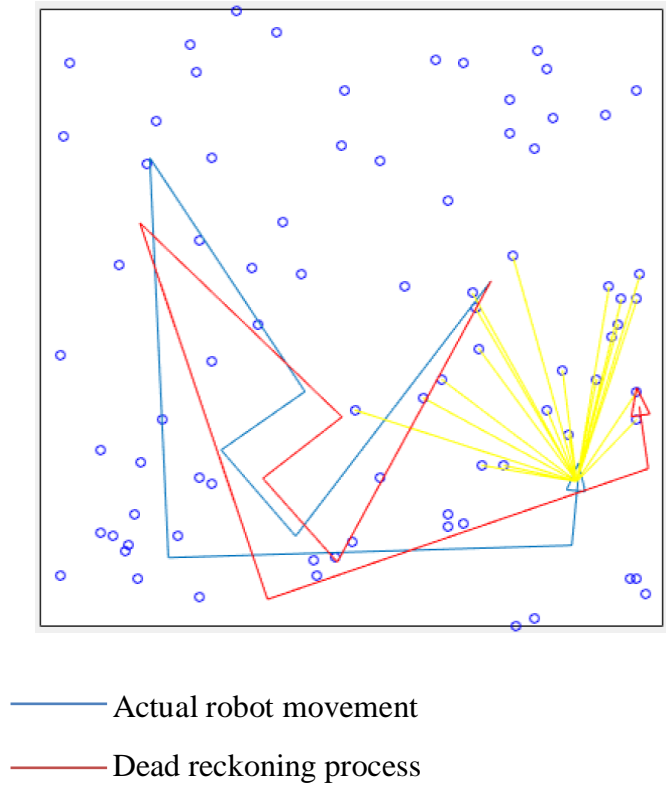

— Actual robot movement
— Dead reckoning process

**Figure 5-3.** Simulation dead reckoning localisation process

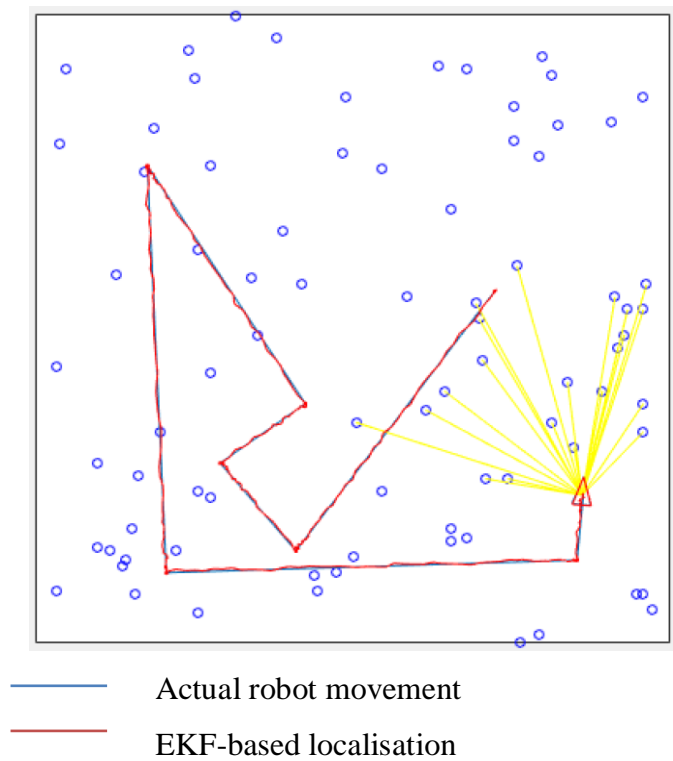

— Actual robot movement
— EKF-based localisation

**Figure 5-4.** Simulation of EKF localisation based on range only measurement

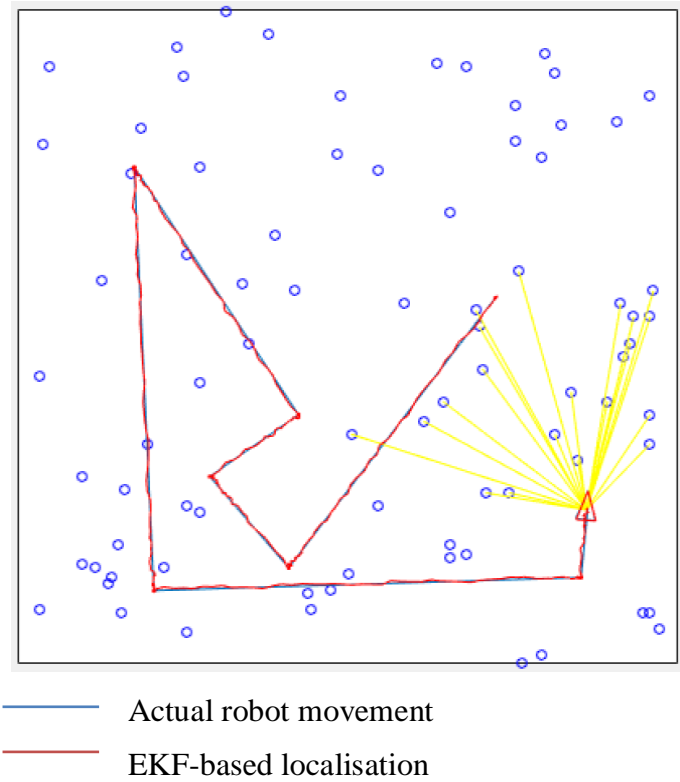

— Actual robot movement
— EKF-based localisation

**Figure 5-5.** Simulation of EKF localisation based on bearing only measurement

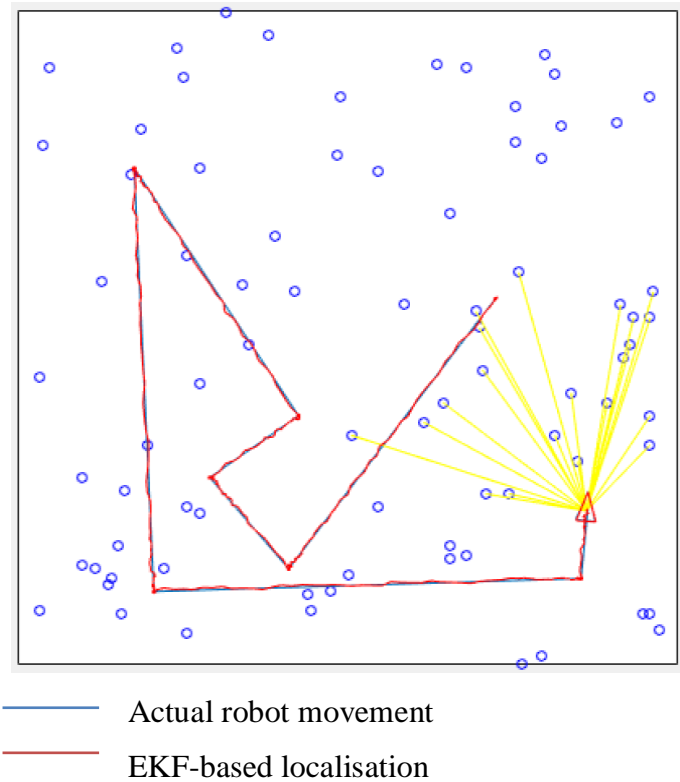

— Actual robot movement
— EKF-based localisation

**Figure 5-6.** Simulation of EKF localisation based on range &bearing measurement

The performance of this EKF-based localisation operation also can be determined from the errors between estimated states and the real robot states during the operation. Figure 7, 8 and 9 illustrate the recorded error in each estimated states (x, y and θ), during the whole operation in various different EKF localisation processes in comparison to dead reckoning process. Table 2 summarizes the root mean square error (MSE) of each operation. These figures and the table also confirm that the EKF-based localisation has proved a successful autonomous operation.

**Table 5-2.** Root Mean Squared Errors of robot localisation processes

| **Dead Reckoning Operation** | | |
|---|---|---|
| Root MSE on x | 9.2721 | meter |
| Root MSE on y | 10.9701 | meter |
| Root MSE on theta | 7.2083 | degree |
| **EKF localisation based on range only measurement** | | |
| Root MSE on x | 0.3683 | meter |
| Root MSE on y | 0.4124 | meter |
| Root MSE on theta | 6.2796 | degree |
| **EKF localisation based on bearing only measurement** | | |
| Root MSE on x | 0.3870 | meter |
| Root MSE on y | 0.4535 | meter |
| Root MSE on theta | 0.3742 | degree |
| **EKF localisation based on range and bearing measurement** | | |
| Root MSE on x | 0.3675 | meter |
| Root MSE on y | 0.3535 | meter |
| Root MSE on theta | 0.2416 | degree |

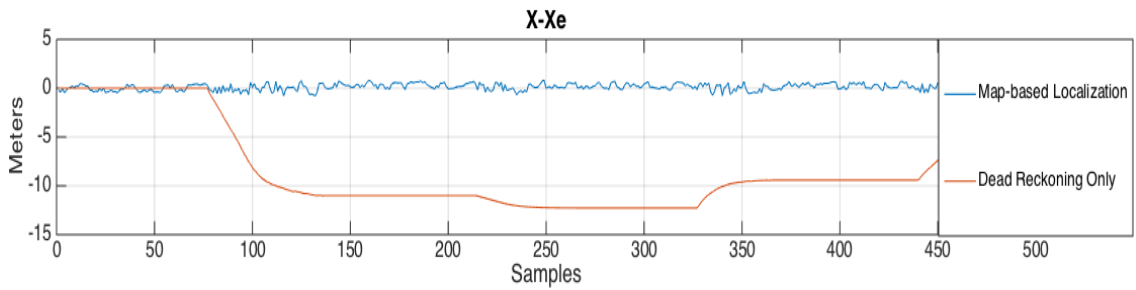

(a)

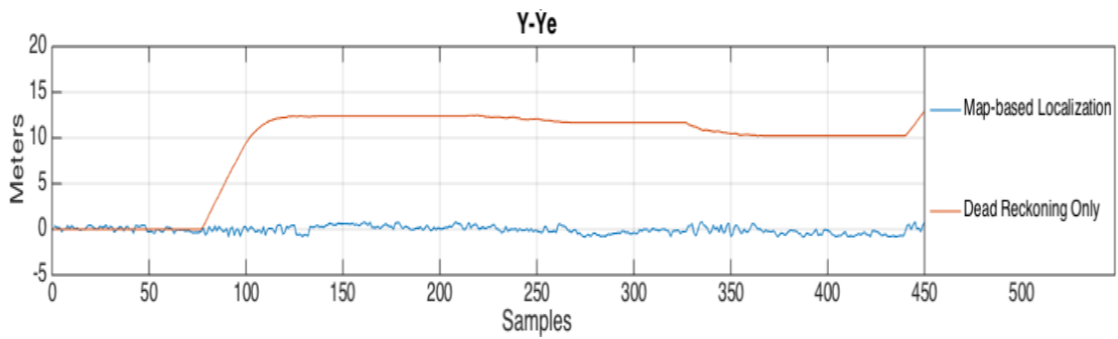

(b)

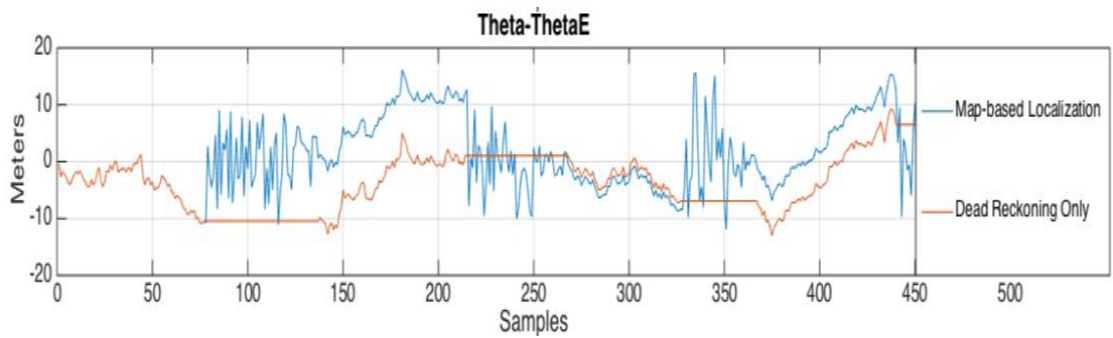

(c)

— EKF-based localization using range only observation process

— Dead reckoning process

**Figure 5-7.** Error in EKF localisation based on range only measurement
(a) Error in x estimation; (b) Error in y estimation; (c) Error in theta estimation

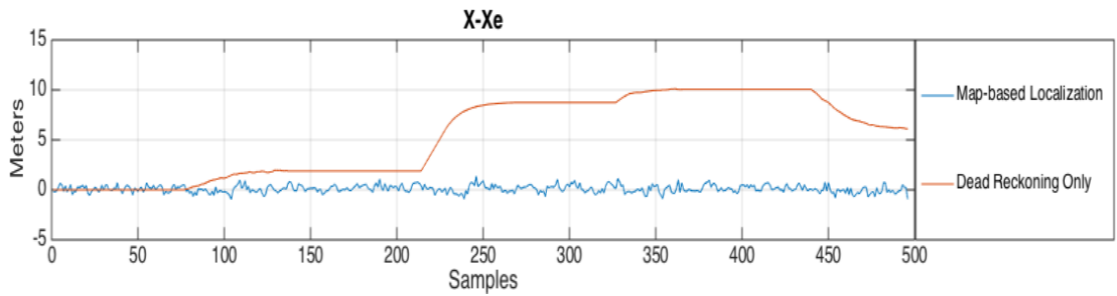

(a)

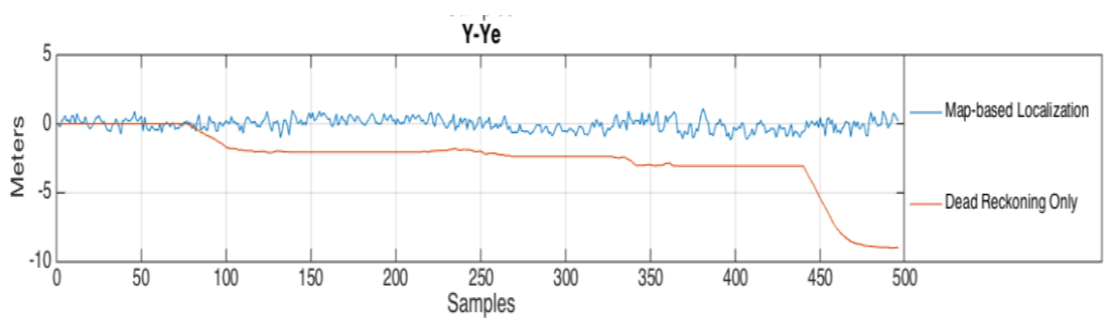

(b)

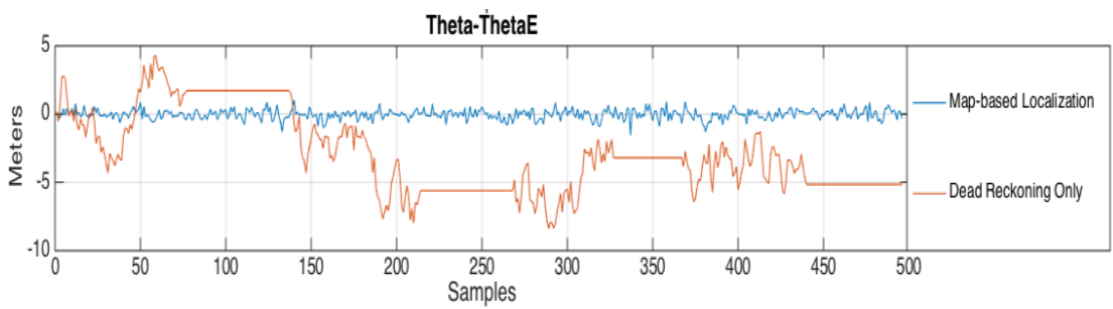

(c)

———— EKF-based localization using bearing only observation process

———— Dead reckoning process

**Figure 5-8.** Error in EKF localisation based on bearing only measurement
(a) Error in x estimation; (b) Error in y estimation; (c) Error in theta estimation

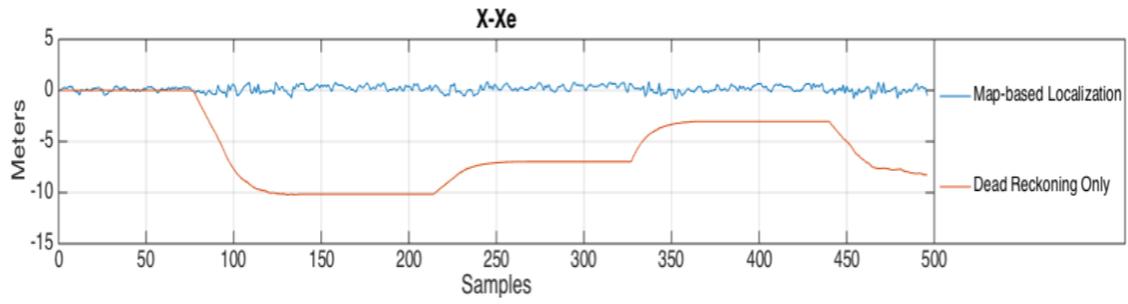

(a)

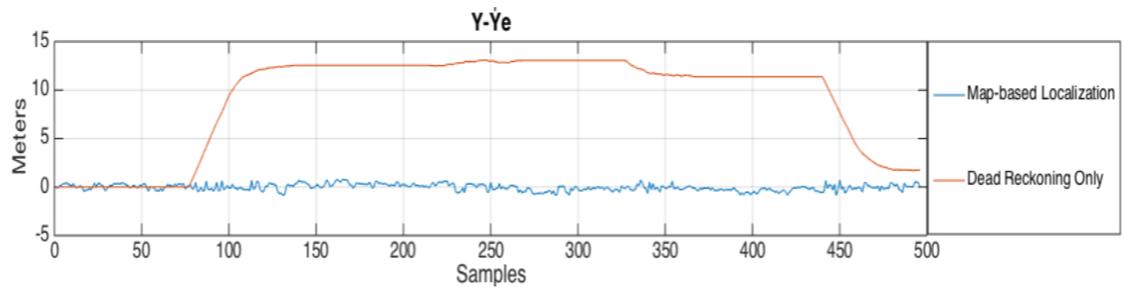

(b)

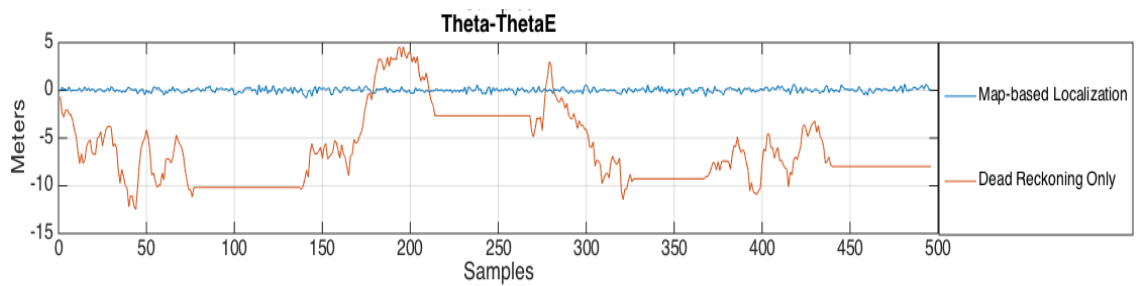

(c)

——— EKF-based localization using range & bearing observation process

——— Dead reckoning process

**Figure 5-9.** Error in EKF localisation based on range &bearing measurement

(a) Error in x estimation; (b) Error in y estimation; (c) Error in theta estimation

# Chapter

# 6. 2D EKF-based SLAM Operation

The overview of SLAM process in this project can be seen on Figure 6-1. SLAM operation consists of several similar processes as in EKF localisation, in which the distinctive feature involves initialisation of landmarks to detect new landmarks and update robot position.

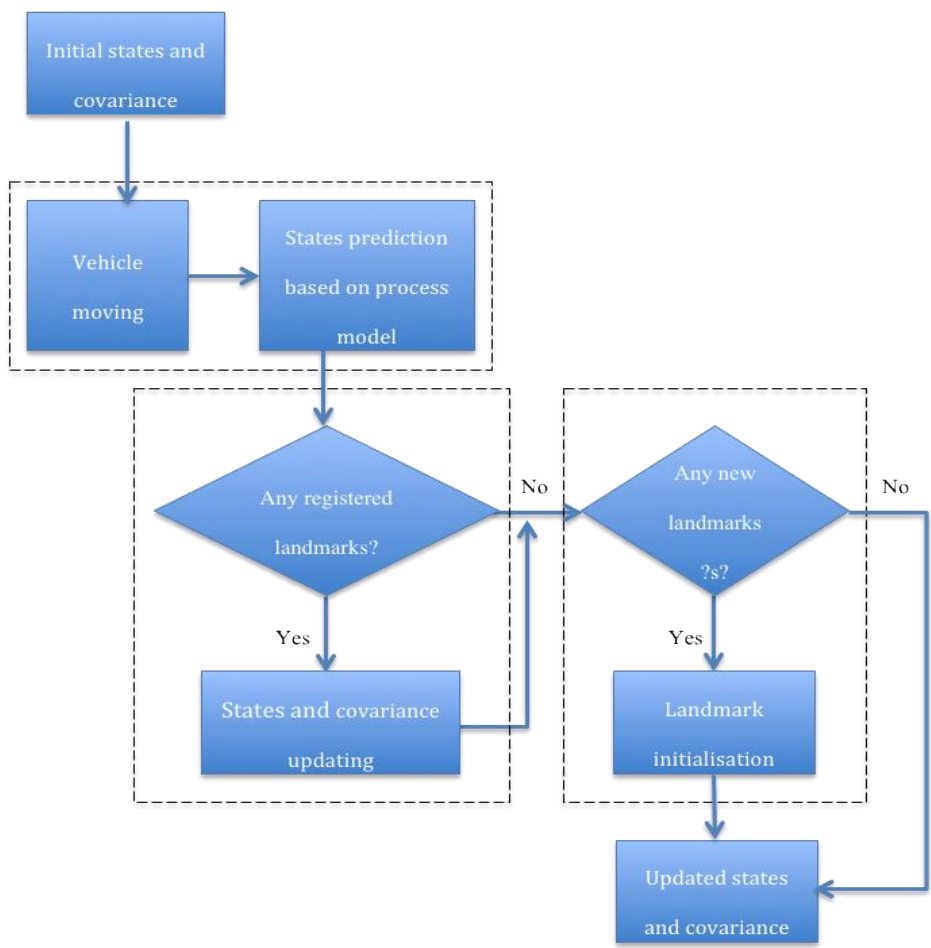

**Figure 6-1**. General operation step for implementation 2D SLAM

The estimated states during the process consist of estimated robot states and estimated landmark states. Landmarks used in this project are point landmarks, which have two-dimensional states(x and y). Equation 6-1 shows the whole estimated states vector in this SLAM operation. In equation 1, $X_v$ correspond to the robot states (x, y, θ) and $X_l$ corresponds to set of landmark states $(Lx_1, Ly_1, Lx_2, Ly_2, ..., Lx_n, Ly_n)$, with n is the number of registered landmark. Similar to the localisation process, the states is estimated by Extended Kalman Filter, in which all the process are modelled by a Gaussian variables including expected value of state vector as a mean and the covariance matrix P. The correspondent of expected state vector and its covariance matrix P can be seen in equations 6-2 and 6-3. Due to new initialisation of landmark process, these two matrixes will expand in size every time the robot detects new observed landmarks.

$$X = \begin{bmatrix} X_v \\ X_l \end{bmatrix} = \begin{bmatrix} x \\ y \\ \theta \\ Lx_1 \\ Ly_1 \\ \vdots \\ Lx_n \\ Ly_n \end{bmatrix} \tag{6-1}$$

$$\overline{X} = \begin{bmatrix} \overline{X_v} \\ \overline{X_l} \end{bmatrix} = \begin{bmatrix} \overline{X_v} \\ \overline{L_1} \\ \vdots \\ \overline{L_1} \end{bmatrix} \tag{6-2}$$

$$P = \begin{bmatrix} P_{X_v,X_v} & P_{X_v,X_l} \\ P_{X_l,X_v} & P_{MX_l,} \end{bmatrix} \tag{6-3}$$

$$P = \begin{bmatrix} P_{X_v,X_v} & P_{X_v,L_1} & \cdot & \cdot & \cdot & P_{X_v,L_n} \\ P_{L_1,X_v} & P_{L_1,L_1} & \cdot & \cdot & \cdot & P_{L_1,L_n} \\ \cdot & \cdot & & & & \cdot \\ \cdot & \cdot & & & & \cdot \\ \cdot & \cdot & & & & \cdot \\ P_{L_n,X_v} & P_{L_n,L_1} & \cdot & \cdot & \cdot & P_{L_n,L_n} \end{bmatrix}$$

## 6.1. General Operation for 2D EKF-based SLAM Process

### 6.1.1. Step 0. Robot Initialization

At the initial time, the initial states only consist of the robot states, and there is still no landmark registered to the map. Therefore, the initial state X = R. Moreover, it is assumed that the initial state of the robot is known exactly, and there is no uncertainty on that belief so that the initial covariance matrix can be considered as P = P₀.

$$\overline{X} = \overline{X_v} = \begin{bmatrix} x \\ y \\ \theta \end{bmatrix} = \begin{bmatrix} 0 \\ 0 \\ 0 \end{bmatrix} \qquad P = \begin{bmatrix} 0 & 0 & 0 \\ 0 & 0 & 0 \\ 0 & 0 & 0 \end{bmatrix} \qquad (6\text{-}4)$$

### 6.1.2. Step 1. Prediction Step

*1. Updating robot states*

When the robot is moving, only the states of the robot, will be affected by the robot movement. Based on the robot movement and its process model, the new estimated of the robot state ($X_v$) is predicted as in equation 6-5 and the landmark state ($X_l$) is predicted as in equation 6-6.

$$\overline{X}_v \leftarrow f_{X_v}(\overline{X}_v, u, \overline{N}) \tag{6-5}$$

$$\overline{X}_l \leftarrow \overline{X}_l \tag{6-6}$$

Equation 6-5 corresponds to the robot process model equation 4-4 based on its previous state ($X_v$) control input (u), and noise model (N). In this process, the noise is modelled as white noise, so that the $\overline{N}$ is equal to zero.

2. *Updating robot covariance*

The updated covariance P based on model prediction in this process is calculated as in equation 6-7, in which $F_x$ is the Jacobian of the state correspond to the process model equation, and Pn is noise covariance of the measurement input control.

$$P \leftarrow F_x P F_x^T + Pn \tag{6-7}$$

As mentioned previously, the robot movement only affects the states of the robot ($\overline{R}$) so the Jacobian matrix to update the covariance matrix P is also only affects the covariance matrix related to the robot states. Therefore, the Jacobean matrix in this process is calculated as in equation 6-8, in which I corresponds to the identity matrix, and 0 corresponds to zero matrices.

$$F_x = \begin{bmatrix} \dfrac{df_R}{dR} & 0 \\ 0 & I \end{bmatrix} \qquad Pn = \begin{bmatrix} \dfrac{df_R}{dN} \\ 0 \end{bmatrix} \tag{6-8}$$

### 6.1.3. Step 2. Updating Process Based on Landmark Observation

While the robot is moving, it observes around that landmark using the laser scanner. The laser scanner measures range and bearing of observable landmarks related to the robot position and orientation. If the observed landmark is already registered to the map, its range and bearing measurement are used to update the states estimation ($\bar{X}$) and also its covariance (P). The measurement process and its corresponded covariance modelled as in equation 6-9 and 6-10 are independent for each landmark (i). The state updating process based on observed landmarks is processed one by one of each landmark.

$$\bar{z} = y_i - h_i(X_v, L_i) \tag{6-9}$$

$$H_X = [H_{X_v} \quad 0 \quad ... \quad 0 \quad H_{L_i} \quad 0 \quad ... \quad 0] \tag{6-10}$$

$$H_{X_v} = \frac{\delta h_i(\bar{X}_v, \bar{L}_i)}{\delta X_v} \; ; \; H_{L_i} = \frac{\delta h_i(\bar{X}_v, \bar{L}_i)}{\delta L_i}$$

Based on this measurement model and its Jacobean (see equation 6-10), the updating state process and updating its covariance are calculated based on a set of equation 6-11 into equation 6-15, in which K is Kalman gain of this updating process.

$$\bar{z} = y_i - h_i(X_v, L_i) \tag{6-11}$$

$$Z = H_X P H_X^T + R \tag{6-12}$$

$$K = PH_X^T Z^{-1} \tag{6-13}$$

$$\overline{X} \leftarrow \overline{X} + K\overline{z} \tag{6-14}$$

$$P \leftarrow P - KZK^T \tag{6-15}$$

### 6.1.4. Step 3. Landmark Initialization

When the robot observes the landmarks for the first time, the landmarks have not registered yet on the map. In this case, the state of this new landmark including x and y global coordinate is estimated based on its range and bearing measurement corresponding to the robot state R. The estimated states of this new landmark are calculated as the invert of observation function ($h_i(R, L_i)$) so that the function can be written as equation 6-16. The corresponding Jacobean related to the new landmark states, robot states, and the inverse observation function are written as equation 6-17.

$$\overline{L}_{n+1} = g(\overline{X}_v, \overline{y}_{n+1}) \tag{6-16}$$

$$G_{X_v} = \frac{\partial g(\overline{X}_v, \overline{y}_{n+1})}{\partial X_v} \qquad G_{L_{i+1}} = \frac{\partial g(\overline{X}_v, \overline{y}_{n+1})}{\partial L_{i+1}} \tag{6-17}$$

The covariance of this new landmark and cross covariance related to the prior states then is calculated based on equation 6-18 and 6-19.

$$P_{LL} = G_{X_v} P_{LL} G_{X_v}^T + G_{y_{n+1}} X_v G_{y_{n+1}}^T \tag{6-18}$$

$$P_{LX} = G_{X_v} P_{LL} P_{X_v X_l} \tag{6-19}$$

Based on the estimation result, these new landmark states and its covariance are then added into the full state of the robot and map and covariance as in equation 6-20.

$$\overline{X} \leftarrow \begin{bmatrix} \overline{X} \\ L_{n+1} \end{bmatrix} \qquad P \leftarrow \begin{bmatrix} P & P_{LX}^T \\ P_{LX} & P_{LL} \end{bmatrix} \qquad \textbf{(6-20)}$$

## 6.2. Simulation Result

In this project, the implementation 2D EKF based SLAM algorithm was implemented in the simulation in MATLAB. The simulation of SLAM operation was performed using the same simulation framework used for EKF-based localisation simulation. Figure 6-2 shows a comparison of the estimating robot location during traveling using only dead reckoning method and EKF localisation. As a comparison, Figure 6-3 shows the estimation of robot location based on EKF-based SLAM operation. It clearly can be seen from the simulation that the slam operation provides higher accuracy and reliability for estimating robot location in long-term travelling process.

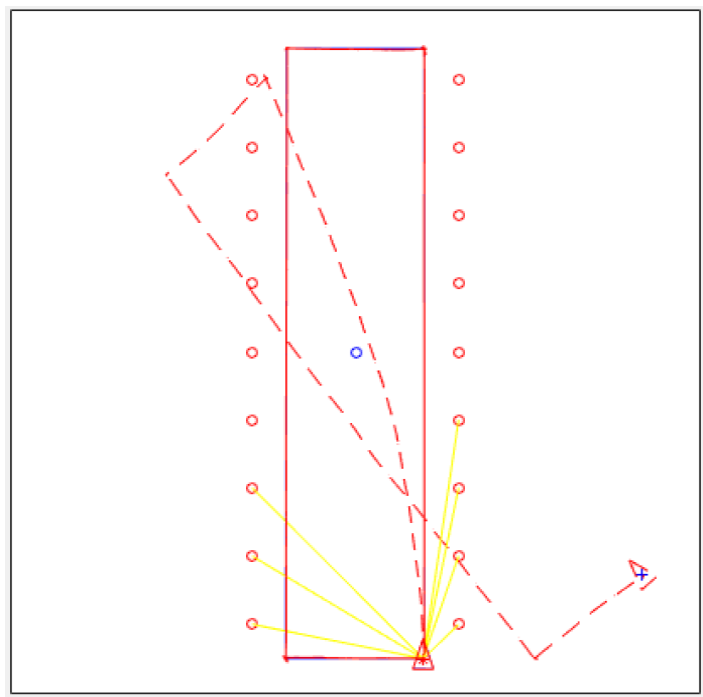

———— Actual robot movement
- - - - - Dead reckoning process
———— EKF-based localization using range & bearing observation

**Figure 6-2.** Simulation dead reckoning and map-based localisation process

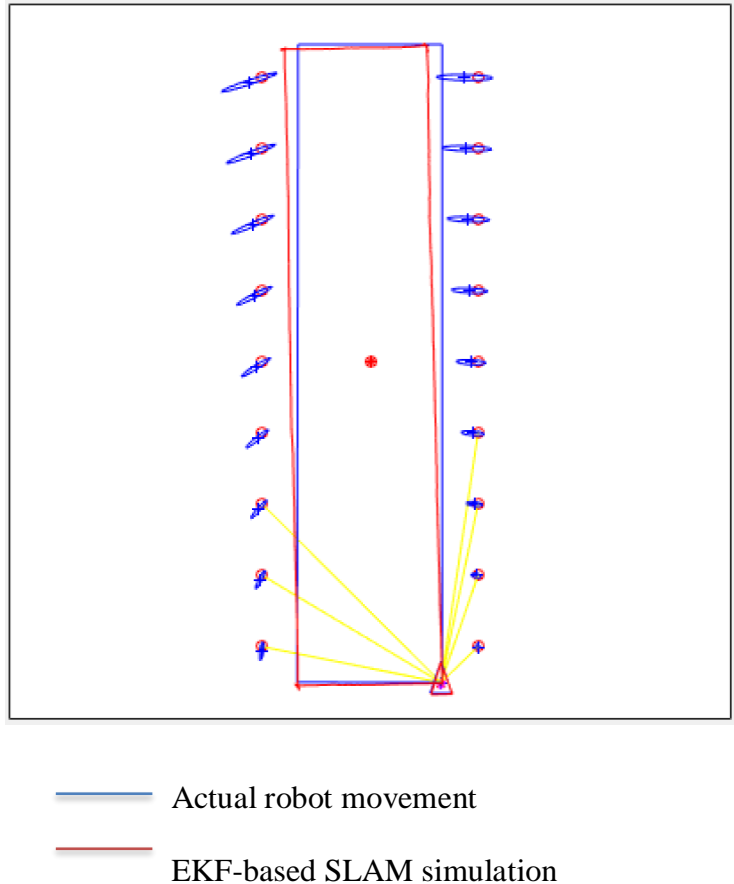

———— Actual robot movement

———— EKF-based SLAM simulation

**Figure 6-3.** Simulation result of EKF-based SLAM

The performance of this SLAM operation can also be evaluated by the state estimation errors. Figure 6-4 illustrates the errors, in states estimation (x, y and θ) in different localisation method. From the figure, it can be clearly seen that the errors in states estimation produced by dead reckoning technique are unbounded accumulated during the iteration process. In contrast, the state estimation produced by EKF-based localisation and EKF-based SLAM indicates fewer errors than dead reckoning method in long-term travelling or iteration process. The calculated roots Mean Squared Errors (MSE) of each operation is summarised on the table 6-2.

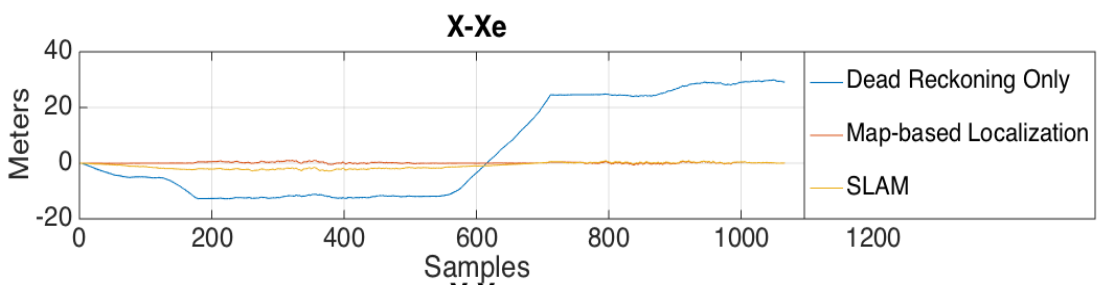

(a)

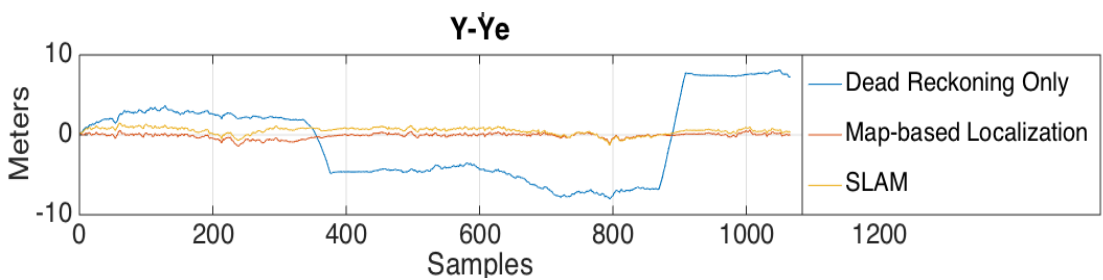

(b)

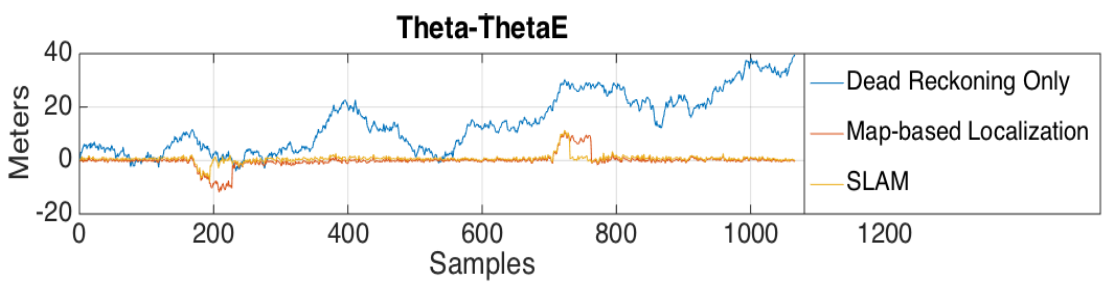

(c)

———— EKF-based localization using range & bearing observation process

———— EKF-based SLAM process

———— Dead reckoning process

**Figure 6-4.** State estimation error in 2D SLAM simulation process

(a) Error in x estimation; (b) Error in y estimation; (c) Error in theta estimation

**Table 6-1.** Root Mean Squared Errors in SLAM process

| **Dead Reckoning Operation** | | |
|---|---|---|
| Root MSE on x | 23.9261 | meter |
| Root MSE on y | 12.6307 | meter |
| Root MSE on theta | 21.1205 | degree |
| **EKF localisation based on range and bearing measurement** | | |
| Root MSE on x | 0.2443 | meter |
| Root MSE on y | 0.2047 | meter |
| Root MSE on theta | 1.7830 | degree |
| **SLAM Operation** | | |
| Root MSE on x | 3.1744 | meter |
| Root MSE on y | 0.6159 | meter |
| Root MSE on theta | 1.7724 | degree |

Another indicator of performance in this SLAM operation is also can be seen in figure 6-5 and figure 6-6. Figure 6-5 shows the comparison of covariance of robot states in these three different methods. Similar to state estimation performances, the robot states covariance is amplified unboundedly during the iteration process. On the other hand, as it can be seen clearly from the figure 6-5b, the robot states covariance is bounded during the iteration in EKF-based localisation and EKF-based SLAM operation. Based on these performance evaluations, it can be concluded that the EKF-based SLAM operation can prevent the robot from being lost during the travelling. In other words, accumulated error occur in dead reckoning process can be minimized using this EKF-based SLAM operation, to achieve more reliable and more precise state estimation of the robot during long-term travel operation.

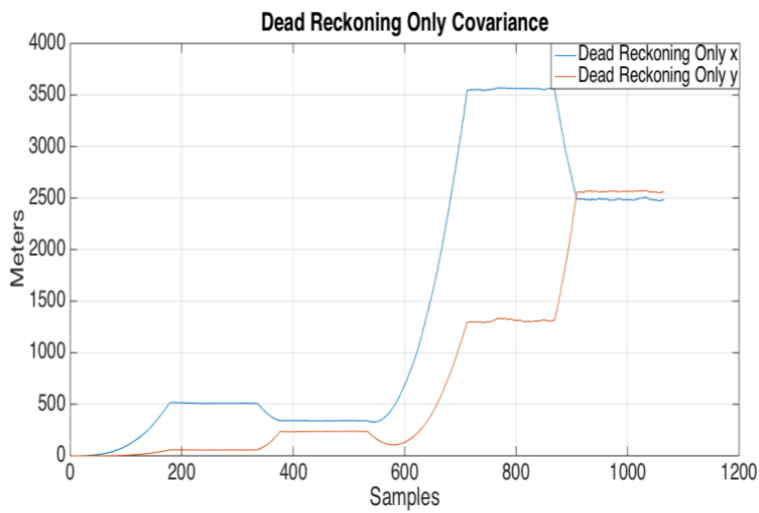

(a)

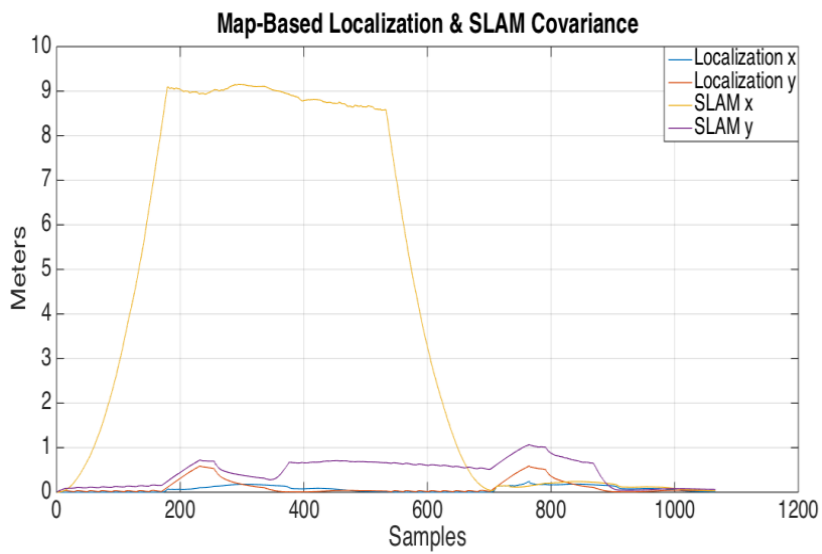

(b)

**Figure 6-5**. Robot's state covariance result in SLAM simulation

The distinctive feature in EKF-based SLAM compared to EKF-based robot localisation is that the landmarks are estimated simultaneously while the robot is travelling. The estimation qualities of the landmarks are updated during the iteration. Figure 6-6 illustrates the graph of the quality of landmarks estimations indicated by its covariance during the SLAM operation. From figure 6-6, it can be clearly seen that the landmarks detected at the first time (i.e. map) has poor quality. Over a period of time, the estimation quality would improve while the robot iterates the operation. Therefore, it can be concluded that the EKF-based SLAM in this simulation performed properly.

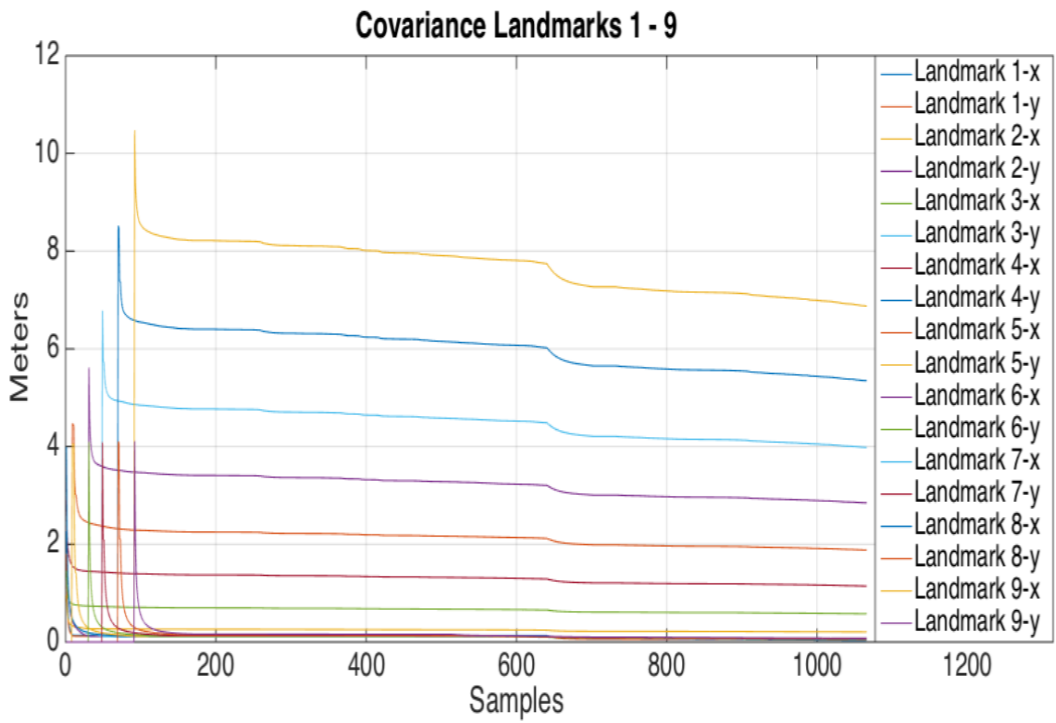

(a)

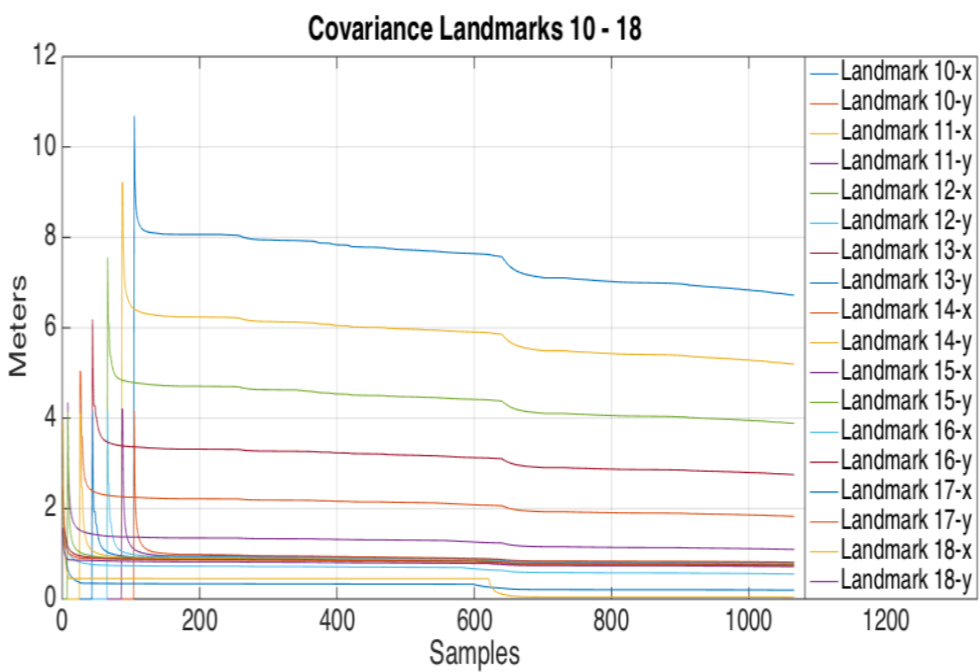

(b)

**Figure 6-6**. Landmark covariance result in SLAM simulation

# Chapter

# 7. Practical Implementation

In this chapter, practical implementations of SLAM operation that has been simulated and evaluated in the previous chapter will be demonstrated. In this case, real data from an Unmanned Ground Vehicle (UGV) will be used to perform SLAM operation in the real environment. In practical case, several processes need to be added and combined with SLAM operation as it is demonstrated in chapter 6, including reading data sensors, estimating and removing bias measurement, landmark extraction, pre-filtering extracted landmark, and data association.

## 7.1. Accessing Measurement Data

In this project several data set has been recorded using UGV platform that can be seen in the figure 7-1. This UGV platform is equipped by odometry sensor, IMU sensor and laser scanner sensor. Set of data produced by these equipped sensor consist of speed data, gyros data, accelerometer, magnetometer and laser scanner range measurement data. Table 7-1 shows the list of properties of the sensors used in this project. For purpose in this project, gyros measurement, speed measurement and laser measurement will be exploited to estimate robot location during the travelling based on EKF-based SLAM operation. The data from GPS receiver will be also utilized to show the comparison between robot localisation based on landmark observation, and robot localisation based on GPS.

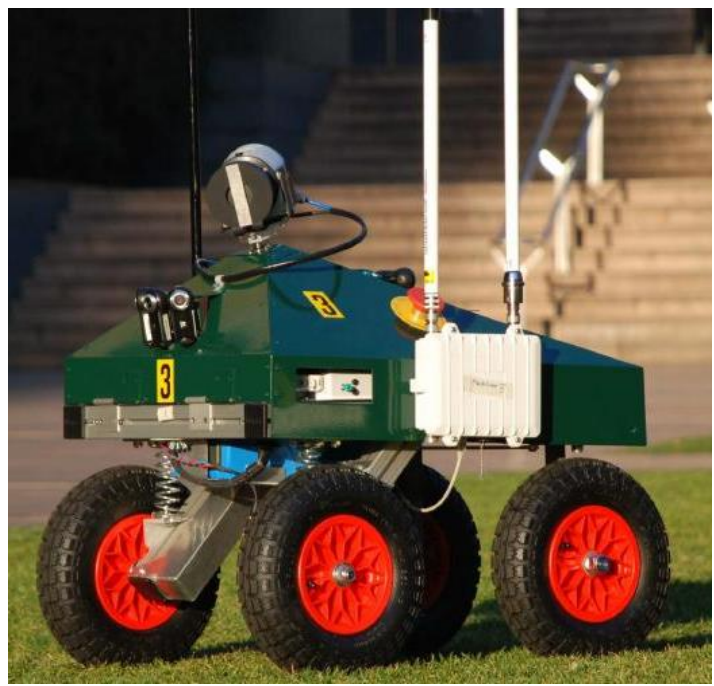

**Figure 7-1.** The UGV platform that used to produce dataset used in this project

**Table 7-1. List of sensors equipped in the UGV**

| Sensor | Description | Value |
|---|---|---|
| **IMU** | Type | MicroStrain |
|  | Sampling rate | 200 Hz |
| **Speed Encoder** | Type | DMC |
|  | Sampling rate | 70Hz |
| **Laser Scanner** | Type | LMS200 |
|  | Sampling rate | 7 Hz |

### 7.1.1. Speed measurement data

Figure 7-2 shows one of the data set of speed measurement of the UGV while it is travelling around UNSW campus.

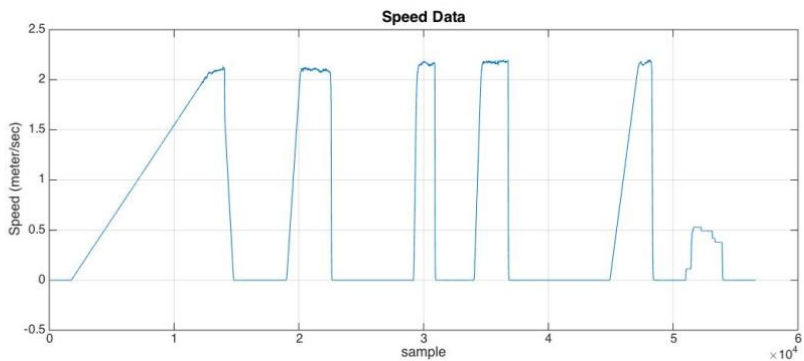

**Figure 7-2.** Speed measurement obtained from the UGV

### 7.1.2. Gyros measurement data

Figure 7-3 shows one of the data set of angular rate measurement on x, y and z axis (i.e. gyros) of the UGV while it is travelling around UNSW campus.

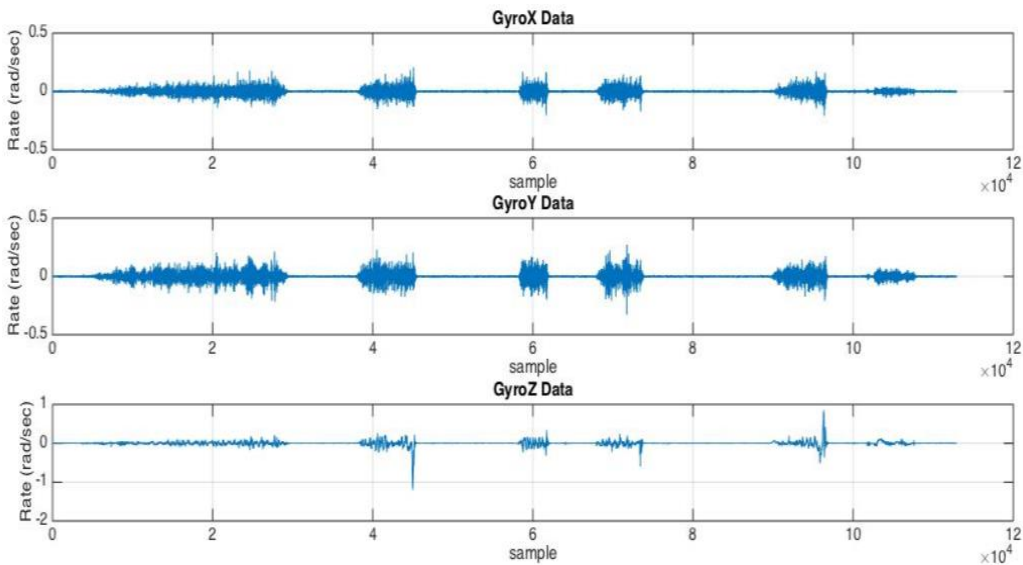

**Figure 7-3.** Gyros measurement obtained from UGV

### 7.1.3. Bias estimation

In some cases, the measured data from the sensors are polluted by systematically errors or known as bias. It can be identify in the first several second of the operation, the robot is not moving. In this case the data measurement suppose to be zero, but in reality, this mean of measurement result in this period is not zero. Figure 7-4 shows an example of gyro measurement polluted by noise.

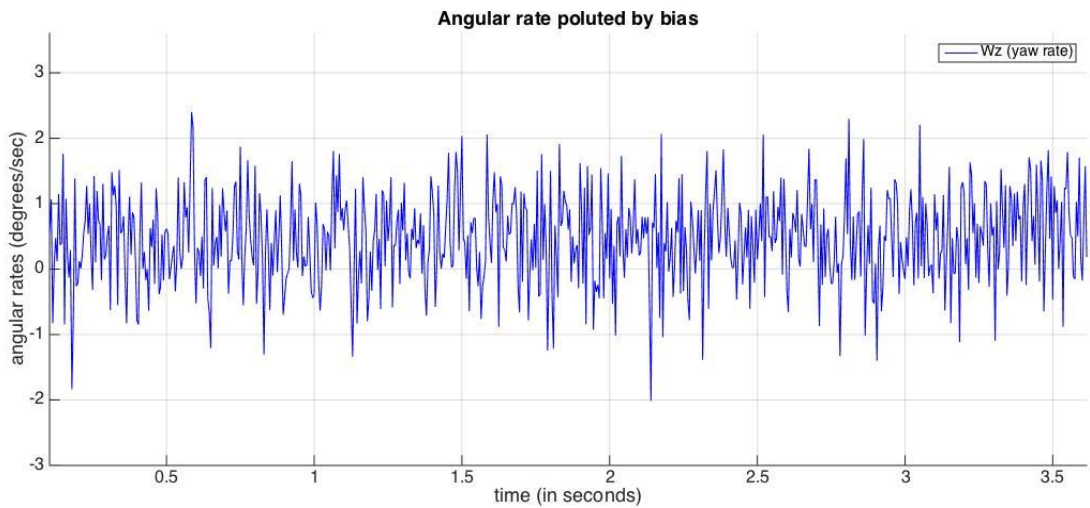

**Figure 7-4.** An example of gyro measurement polluted by noise

In this project, the bias in the measurement of speed and gyros are assumed to be constant. Thus the bias of these measurement data are estimated by calculating the mean values of the measurement data in the first 5 second period while the UGV is still not moving yet. The estimated bias then removed from the entire measurement data. Figure 7-5 shows the measurement data same as in figure 7-4 after removing estimated biases.

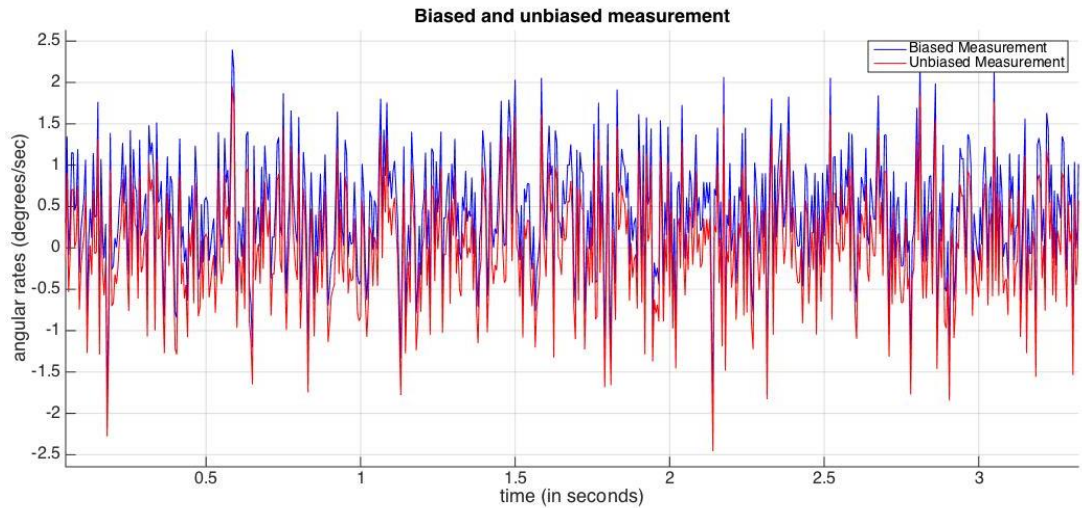

**Figure 7-5.** Comparison biased and unbiased measurement

According to the figure 7-5 the difference between biased and unbiased measurement seems not significant. However, when this data is used in integral calculation, for instance estimating the heading of the robot based on angular rate measurement, this error will be unboundedly escalated. Therefore, this bias estimation process is one of the important parts in this project. Figure 7-6 shows the escalated error causing by bias measurement in estimating robot heading based on angular rate measurement. From the figure, it can be seen clearly the error caused by bias measurement leads to significant error in the estimation process by integral calculation. This error will continuously escalated during the estimation process operating.

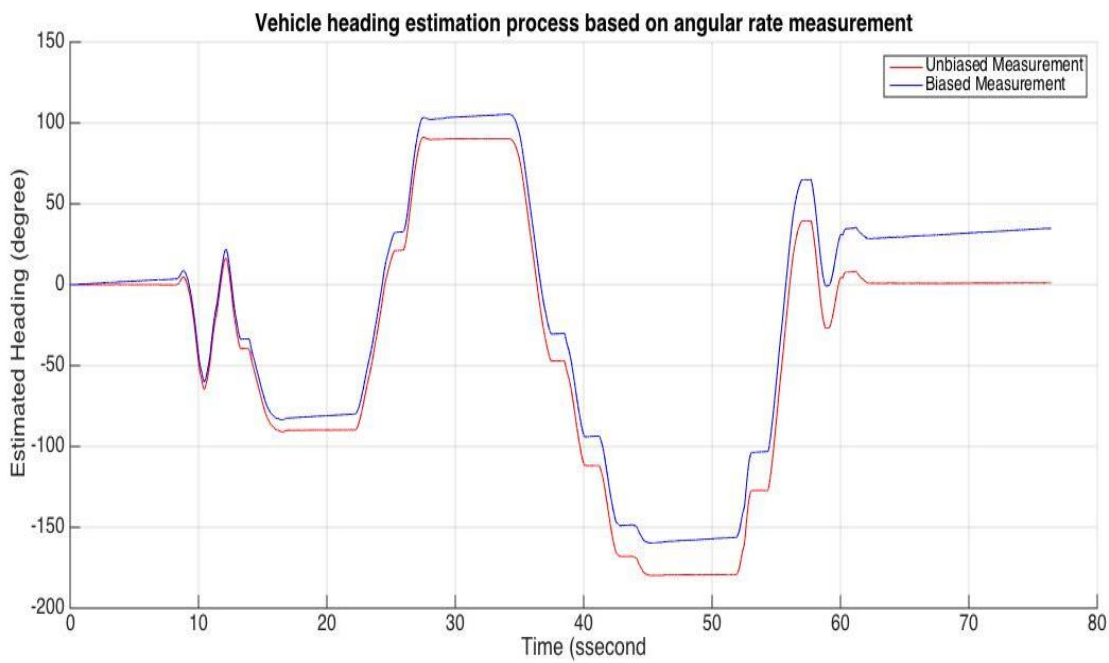

**Figure 7-6.** Drift caused by biased measurement in integral calculation

### 7.1.4. Laser scanner data

The laser scanner sensors used in this project produce 361 range data in every cycle, which cover 180 degree scanning angel. It means that the bearing of the detected point object can be estimated with 0.5-degree resolution. Figure 7-7 shows the plot of range data obtained by laser scanner with respect to the scanning angle (i.e. bearing). Based on these scanning angles and range measurement, the detected point objects can be mapped into Cartesian coordinate in the UGV coordinate frame. Figure 7-8 shows the detected point objects produced by lased scanner, mapped in Cartesian robot coordinate.

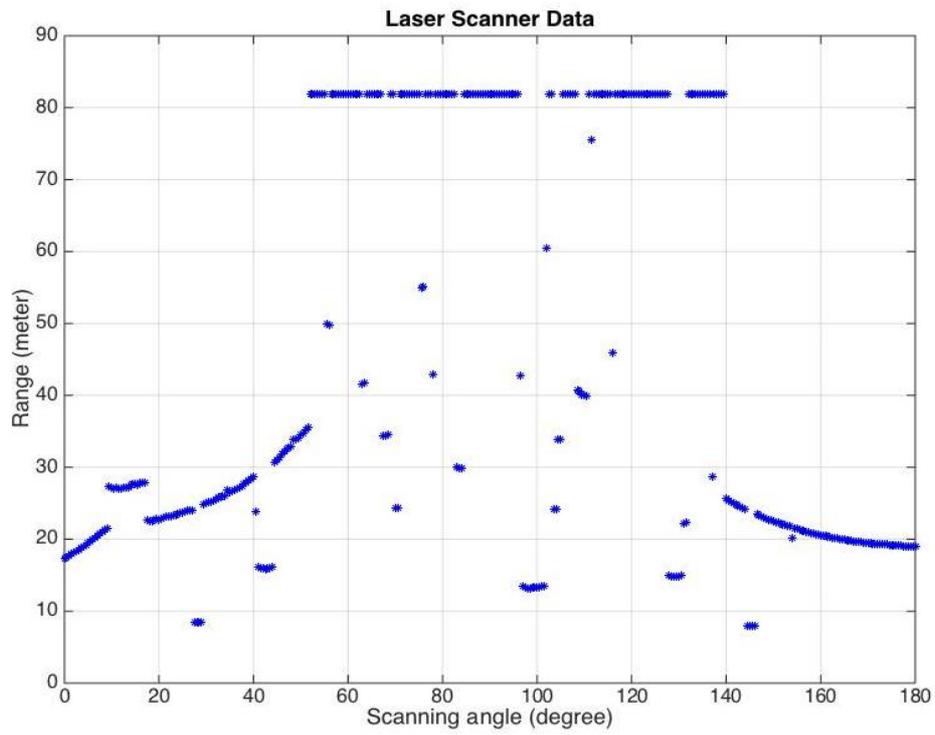

**Figure 7-7.** Data provided by laser scanner sensor

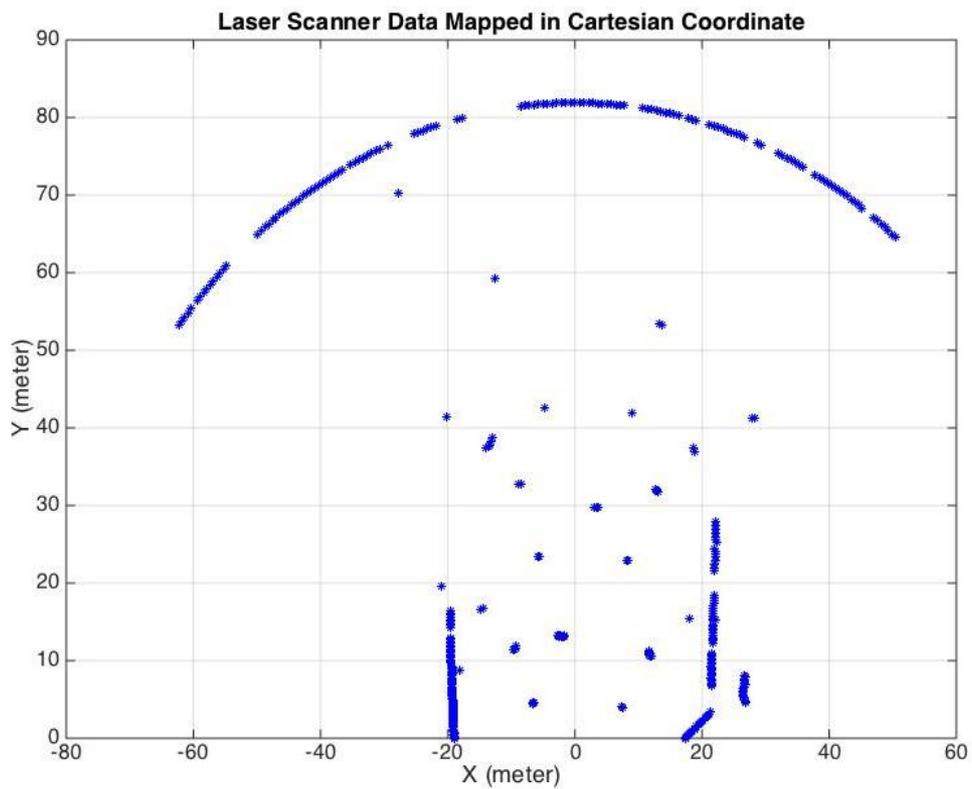

**Figure 7-8.** Laser scanner data mapped in Cartesian coordinate in robot coordinate frame

## 7.2. Landmark extraction

### 7.2.1. Landmark extraction criteria

A landmark is a feature or objects that can be easily distinguish and re-observed from the rest of the environment. It will be used for updating process in SLAM operation. Thus, well identification on good landmarks is really important to support SLAM operation. As it can be seen in the figure 7-7 and 7-8, laser scanner measure range point by point in 180-degree workspace. This data then analysed to extract object of interest within these points. The method used for extracting object from the point data in this project is using the fact that points that are belong to an object have range and bearing close to each other and they are separated from the other points. Figure 7-9 and 7-10 shows the detected objects extracted from the points from the laser measurement data.

A landmark is identified by the size of the separated objects on the detected point objects. In this project the point objects are categorised as landmarks if it has size between 3 and 8 points. This size is chosen as estimation of the pole size around UNSW, which can be categorize as good landmarks as it can be easily to be detected by the robot. This landmark identification can be expressed as in equation 7-1. Figure 7-11 shows the landmark extraction from the detected point object by laser scanner.

$$O(x,y) : \begin{cases} 3 < n(O) < 8 & \rightarrow O = L_k \quad (landmark) \\ otherwise & \rightarrow O \neq L_k \quad (not\ landmark) \end{cases} \quad (7\text{-}1)$$

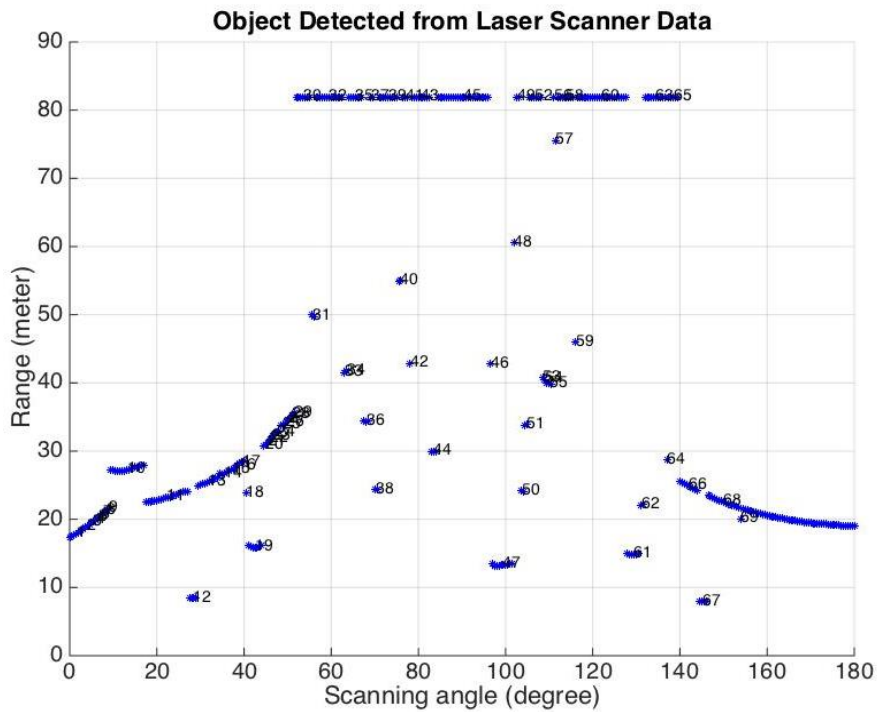

**Figure 7-9.** Object detected from laser scanner data

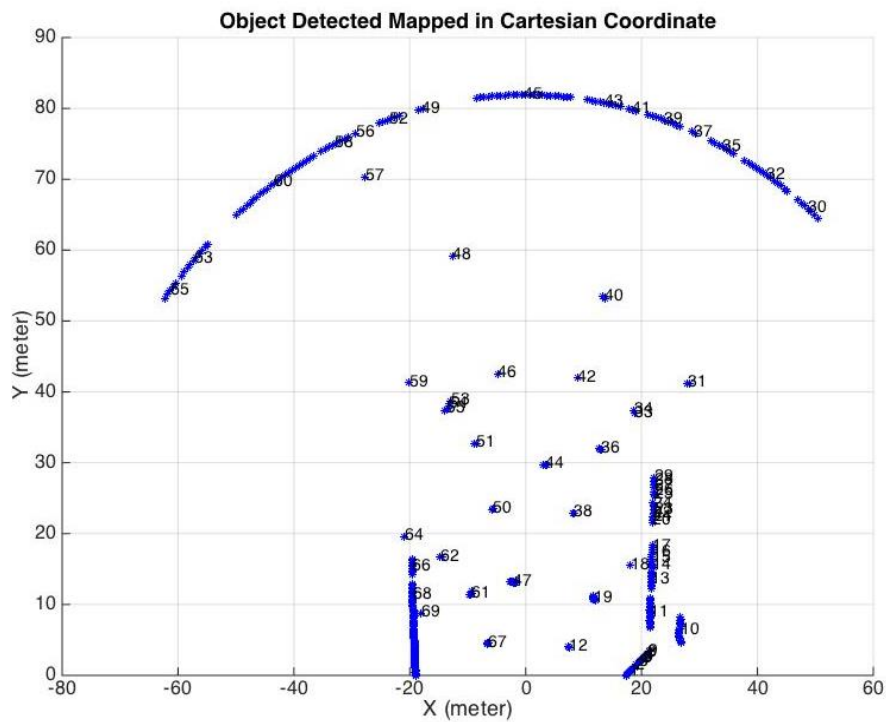

**Figure 7-10.** Object detected mapped into Cartesian coordinate in robot coordinate frame

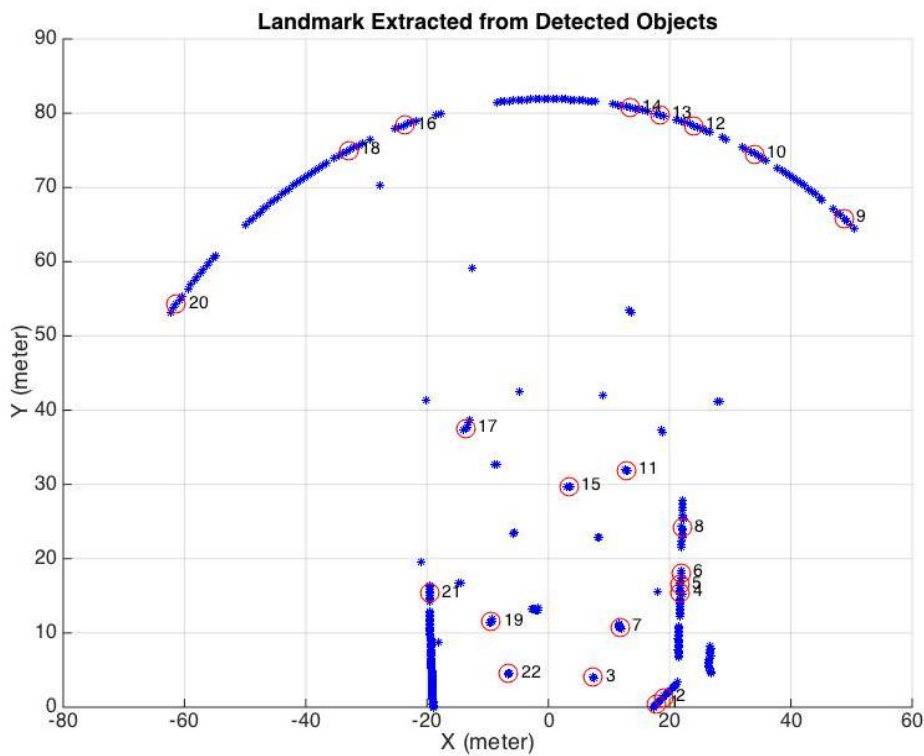

**Figure 7-11.** Landmarks extracted from detected objects

### 7.2.2. Pre-filtering the extracted landmark

As it is mentioned before, landmarks extracted from the measurement data will be used for updating process in this SLAM operation. Having preferable landmarks is critical to produce good estimation results. As it can be seen in the figure 7-11, the extracted landmarks, are containing some inadequate landmarks. For instance, landmarks 9, 10, 12, 13, 14, 16, 18 and 20 are not good to be used as landmarks, since they are not belong to the real objects but its belong to range limit of the laser scanner sensor. Another poor landmarks that can be seen from the figure 7-11 are landmarks 1, 2, 4, 5, 6, 8 and 21. Practically, they are part of buildings or any features on there that detected as separated objects. This objects is potentially not re-observable, or even it is observable, it will result poor accuracy in estimating the position of this landmark, which is not good for whole SLAM operation.

A good landmark is categorised if it satisfy the following requirements:

1. Good landmark has to be apart from the others. In this case, a landmark is categorised as a good landmark if it is separated more than 2 meters from the other landmarks.

2. Good landmark has to be statics. In this case, when the landmark is detected for the second time, the position of this landmark should be remaining the same with variation less than 1 meter.

3. Good landmark has to be re-observable. Detected landmark will be categorised as adequate landmark if this landmark is re-observed multiple time in the next following observation processes.

Figure 7-12 shows the extracted landmark after pre-filtering filtering process based on criteria mentioned before.

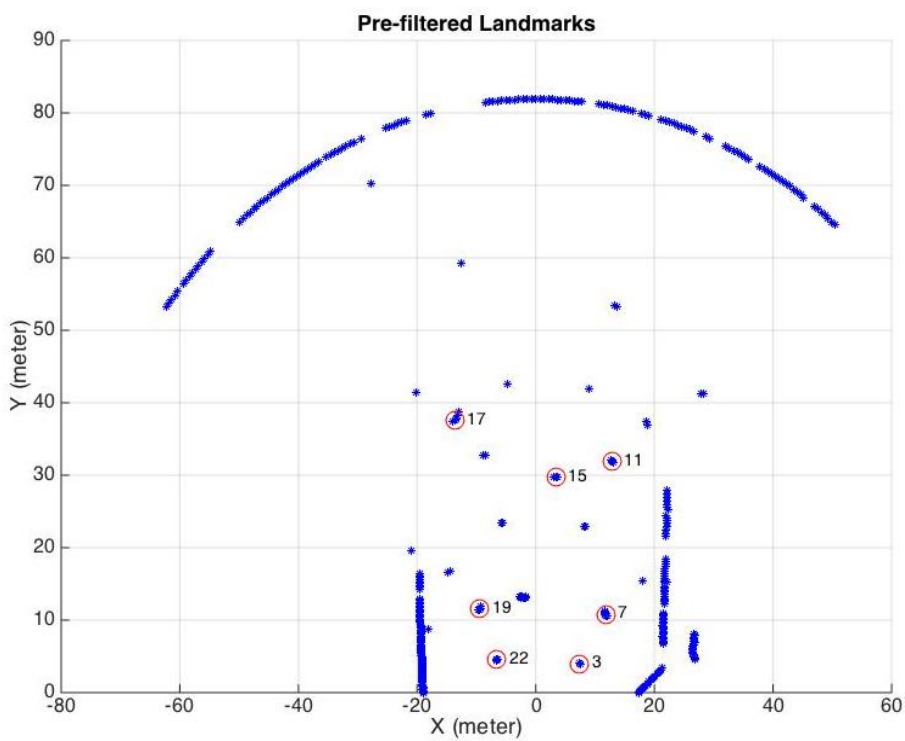

**Figure 7-12.** Pre-filtered Detected Landmarks

## 7.3. Dealing with dynamics objects

In the real implementation, it is often that the UGV have to deal with dynamic objects such as another vehicle moving and pedestrian. Sometimes, the features of these dynamics objects are similar to the landmark criteria that have been set. Thus, these dynamics objects are detected as landmark by the robot. When these objects used as landmarks in SLAM operation, it will be such a bad idea. When these landmarks always detected not in the same place, it might cause two problems. First, if the objects are moving slow, these objects might be re-observable and can be associated with the previous observation. The problem with this case is the robot then will update the estimated position in SLAM operation with the measurement result of these landmarks, which in reality are already moving. This situation will lead to bad estimation results. The second problem is, when the objects are moving fast, These objects cannot be associated with the prior observation, since the UGV assume that the objects in the different location. Then, the UGV will register these objects as new landmarks. This situation will escalate the size of the map, which make the operation slower due to the calculation, without significantly improve the estimation process in SLAM operation. Thus, dealing with dynamics objects is another important part of implementing SLAM in the real environment. Figure 7-13 shows the dynamics objects, which have features similar to the landmark category in this project. Figure 7-13 a and b show the change of these objects in the different time frame.

For dealing with this dynamics objects, a landmark quality parameter, $q$, is addressed. Scoring approach is used to represent the quality of landmarks. In this approach, there are two main sub-operations, called upgrade and degrade. Mathematically this sub-operation can be expressed as in equation 7-1.

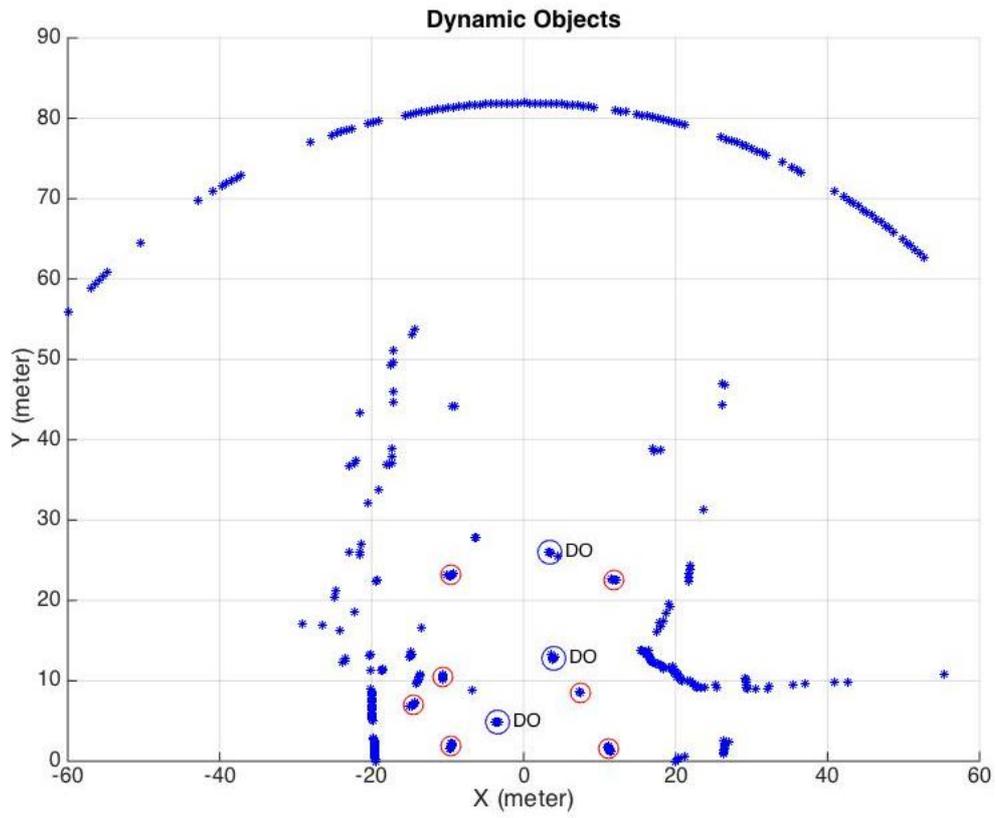

(a)

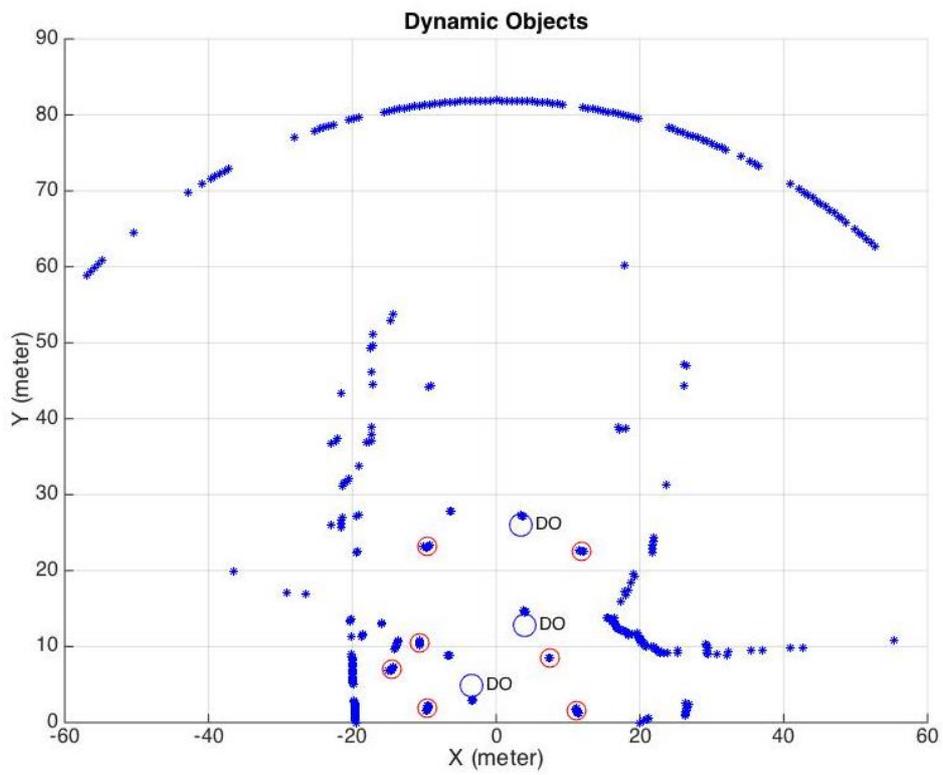

(b)

**Figure 7-13.** Dynamic objects detected as landmarks

$$O(x,y,q) : \begin{cases} O \text{ is reobserved} & \rightarrow q = q + 1 \quad (upgrade) \\ O \text{ is not reobserved} & \rightarrow q = q - 3 \quad (degrade) \end{cases} \quad (7\text{-}2)$$

When the landmark re-observed, in the next iteration the quality of landmarks will increase by one. In contrast, if the landmark is not re-observed in the next iteration, this quality decreases by two. In this project, a landmark is categorised as good landmark if the quality reach 10. This landmark then will be set into database to be used in SLAM operation. On the other hand, a landmark is categorised as poor landmark when the quality reach -20. This landmark then will be clear from the database. These two sub-operations, 'set' and 'clear', can be express as in equation 7-2.

$$O(x,y,q) : \begin{cases} q > 10 \quad (set) \\ q < 15 \quad (clear) \end{cases} \quad (7\text{-}3)$$

## 7.4. Data Association

When the robot performs re-observation operation, detected landmark in this observation should be matched with the landmark produced by prior observation. This process is called data association. This data association is important, in order the UGV can update the states estimation based on re-observed landmarks. The technique used for this landmark association process is "nearest-neighbour approach" [**32**]. One way to perform this technique is by calculating the Euclidean distance. In this technique, the current landmarks observed associate with the prior landmarks that have been registered into database with the closest Euclidean distance with this detected landmarks. This approach mathematically can be expressed as in equation 7-3. In this equation, element $N_i$ , which represent new observed landmarks is associated to element $P_i$, which represents the prior landmarks. The maximum distance for this process is set as 30 cm.

$$B_j = \underset{B_k}{Argmin}\big(dist(A_i, B_k)\big) \tag{7-4}$$

$$A_i = \underset{A_k}{Argmin}\big(dist(A_k, B_j)\big)$$

$$distance(A_i, B_j) = dmax$$

Figure 7-14 and 7-15 shows the observed landmark in two different time frames. Data association process is performed in each observation process. The figures 7-15 show the landmark observed after 30 times observation (about 4.5 sec), associated to the landmarks from the prior observation. From the figure 7-15 it can be seen that the observed landmarks that are already detected in the prior observation are successfully associated, while the new observed object is identified as a new landmark.

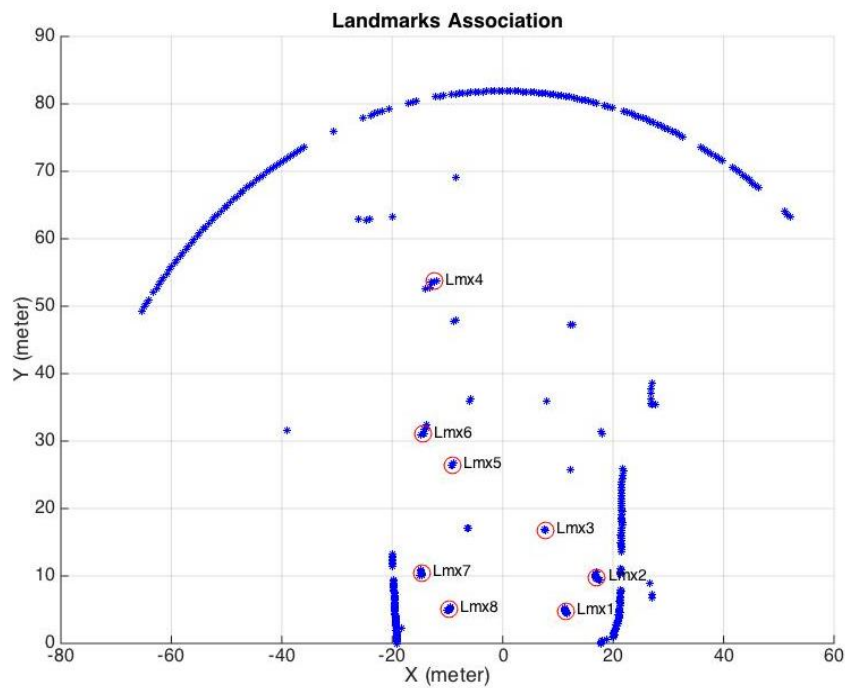

**Figure 7-14.** Landmark detected in the prior observation

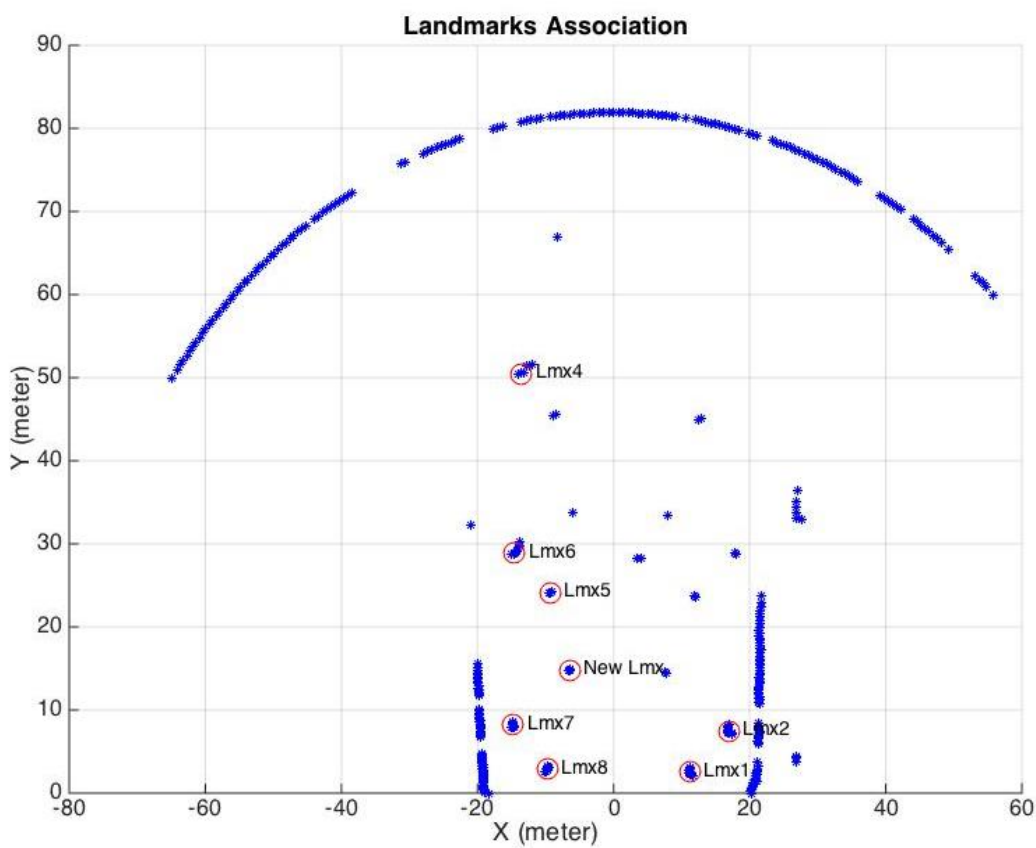

(b)

**Figure 7-15.** Landmark detected in the new observation associated to the prior observation, after 30 iteration while the UGV moving

## 7.5. Experiment Result

### 7.5.1. SLAM operation result

Using data set produced by the UGV's sensors and performing these sub-operations mentioned at the previous sub-chapters, the SLAM operation then could be implemented for this practical implementation. Every step in SLAM operation is performed sequent same as the operation in the SLAM simulation in the previous chapter. In this practical implementation, the control input data, $u = [v\ \omega]'$, used in prediction step, is obtained from speed and gyro-Z (i.e. angular rate in z axis) measurement, after removing the bias as it explained in the previous sub-chapter. Similarly, in the updating process, the landmarks observations are come from the real measurement from the laser scanner after pre-filtering process. In this experiment, the state estimation process obtained from EKF-based SLAM operation will be compared to the state estimation process using dead reckoning method.

Figure 7-16, 7-17 and 7-18 show the UGV path estimation, in three different experiment, based on dead reckoning operation, while the UGV is traveling in the main road of UNSW campus. From these figure it can be seen clearly that the UGV path, estimated by dead reckoning operation is drifted while the robot is travelling. As a result, the map building by the laser scanner observation is also distorted.

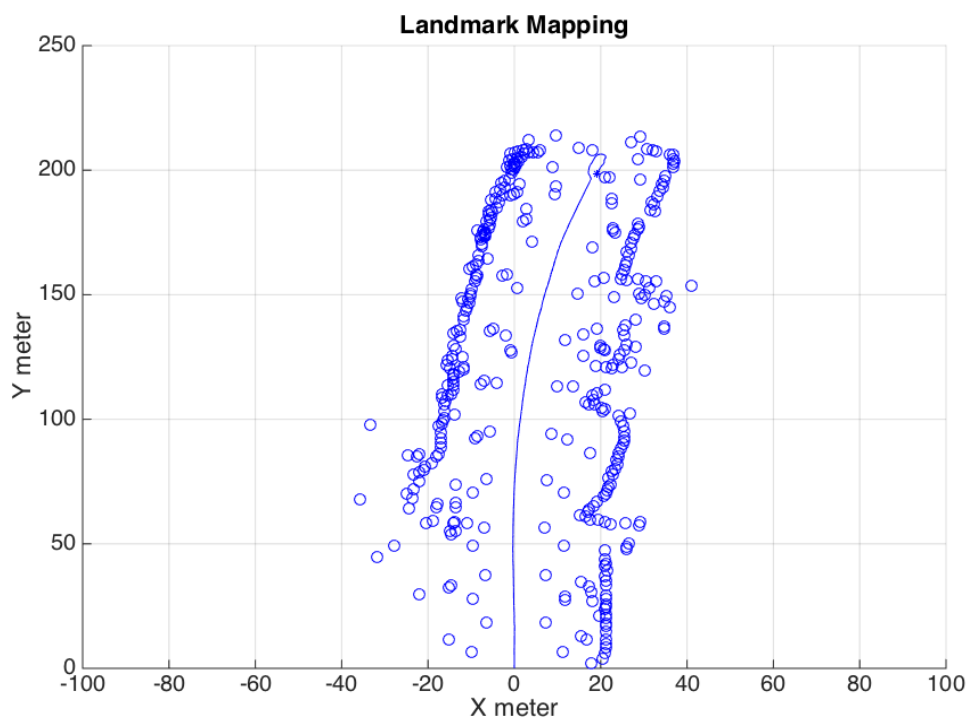

**Figure 7-16.** Robot localisation and mapping using dead reckoning scenario 1

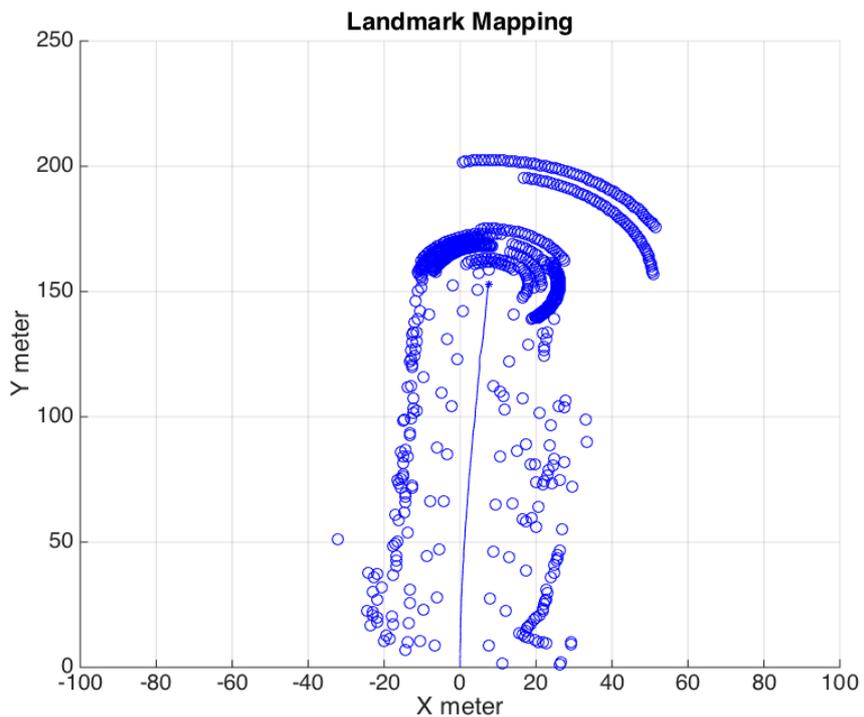

**Figure 7-17.** Robot localisation and mapping using dead reckoning scenario 2

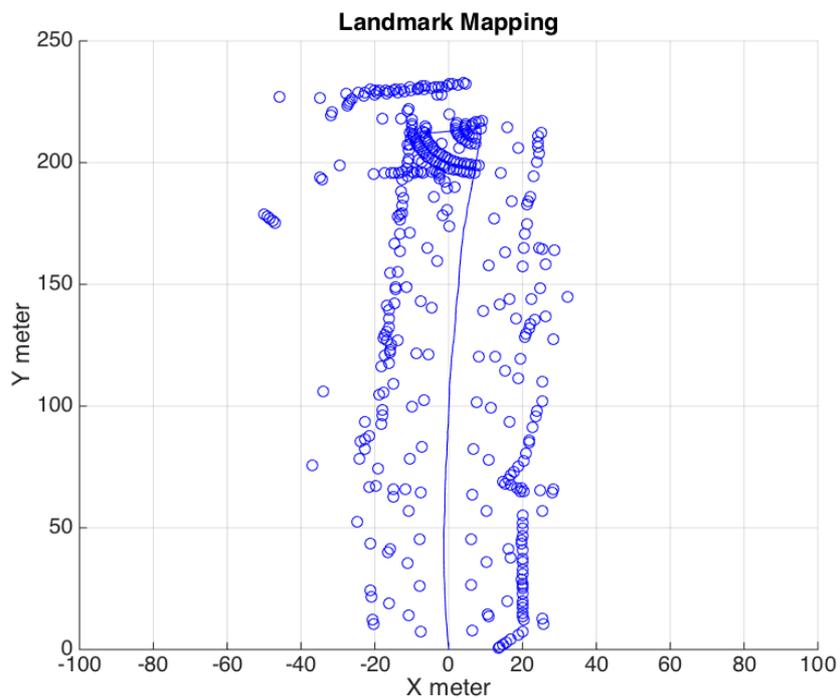

**Figure 7-18.** Robot localisation and mapping using dead reckoning scenario 3

In comparison, figure 7-19, 7-20, and 7-21 demonstrate the UGV path estimation based on EKF-based SLAM operation, without pre-filtering process and landmark quality approach. In this case, poor quality of landmarks, and also moving objects are used in updating process in SLAM operation. From the figures it can be seen briefly that the path estimation and maps resulted from this process are not significantly better than the result of dead reckoning process. The paths estimated are still significantly drifted and the maps produced are distorted. Moreover, the maps generated from this process looks messier than the maps resulted from the dead reckoning, since poor quality of landmarks and also dynamic objects are including on the map. These objects are detected multiple times as new landmarks, which also make the computational time of the process escalated significantly. Figure 7-22 and figure 7-23 shows the comparison of computational time between SLAM operation without pre-filtering landmarks and with pre-filtering landmarks, respectively, in 100 iterations. Based on this comparison it can be seen the significant amplifying of time consuming in these two SLAM process in 100 iterations. The calculation time in SLAM operated without pre-filtering landmarks process is amplified by about 7 times in only 100 iterations. Thus, this will not be practical to be implemented.

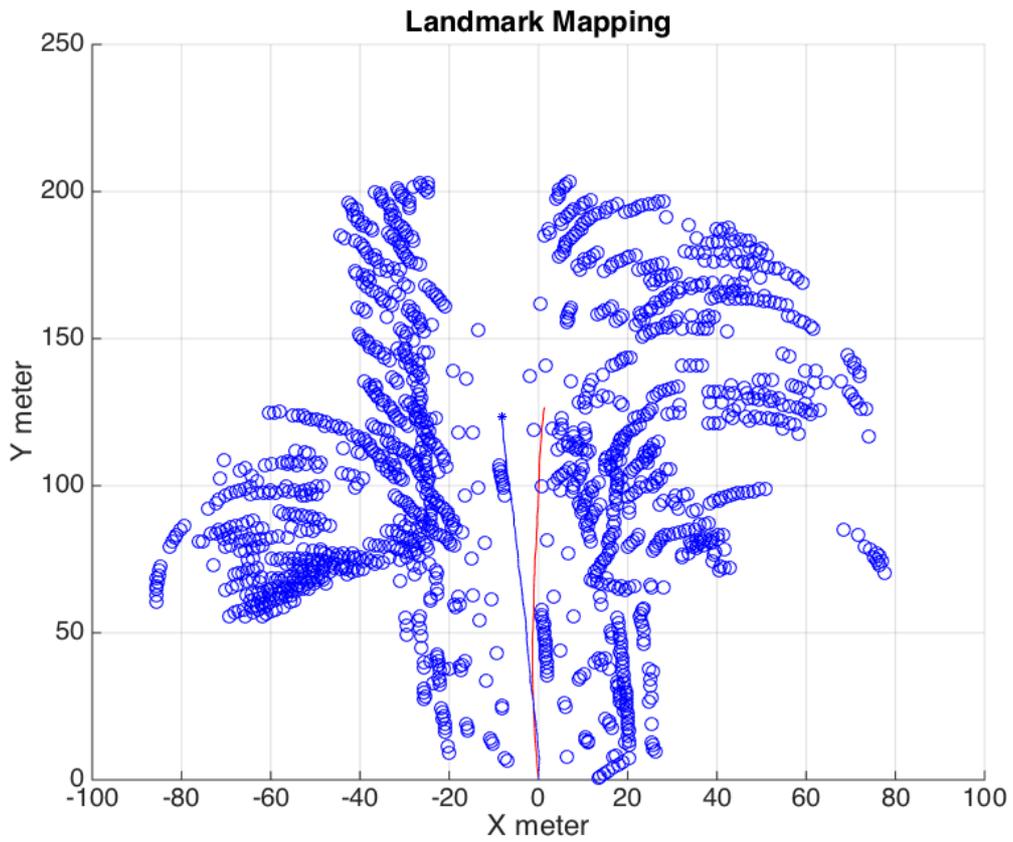

**Figure 7-19.** SLAM operation without prefiltering scenario 1

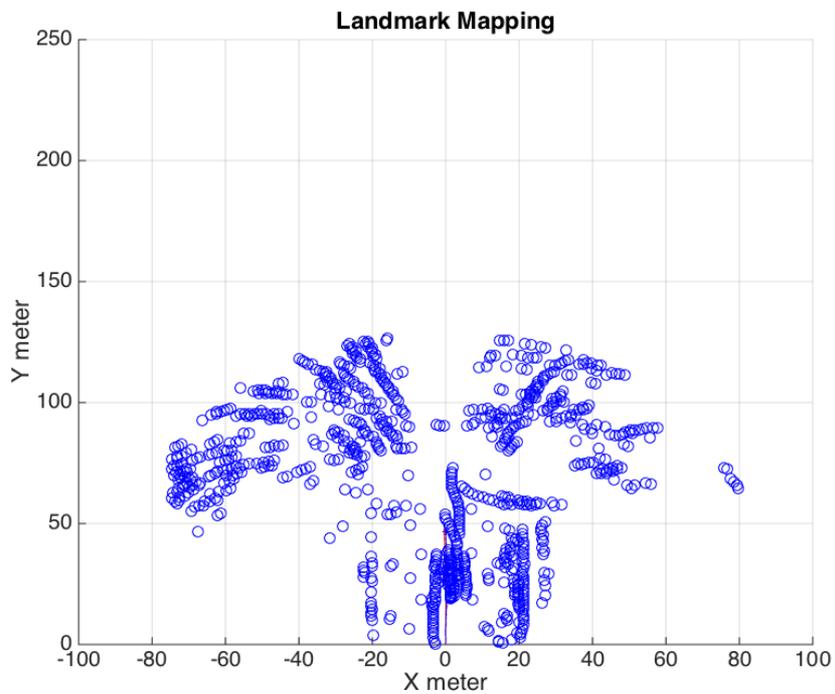

**Figure 7-20.** SLAM operation without pre-filtering scenario 2

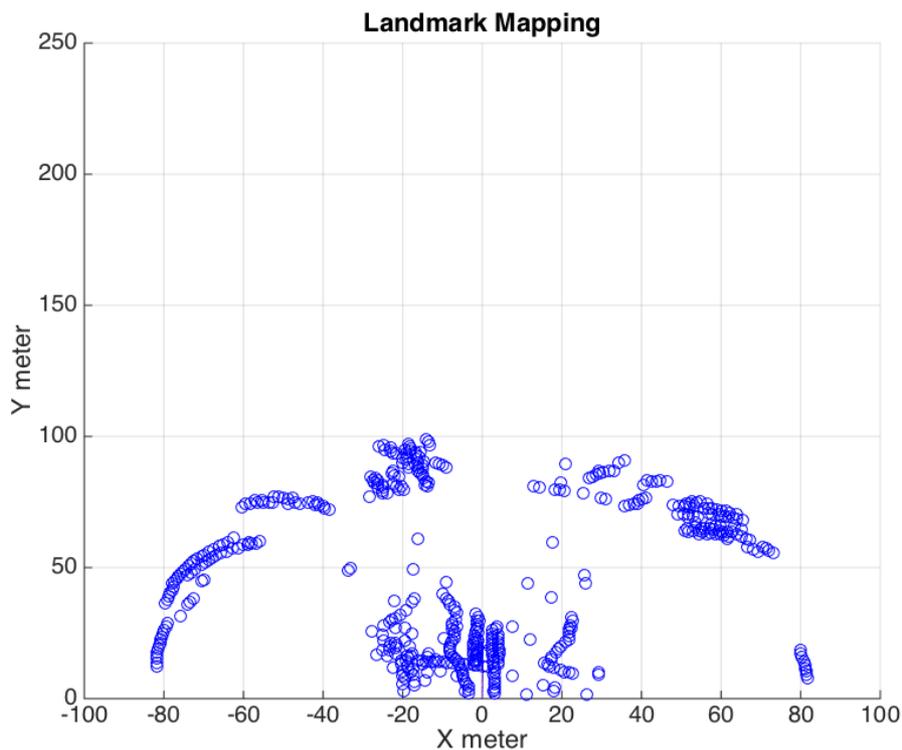

**Figure 7-21.** SLAM operation without pre-filtering scenario 3

| Function Name | Calls | Total Time | Self Time* | Total Time Plot (dark band = self time) |
|---|---|---|---|---|
| MainProgram | 1 | 76.314 s | 2.467 s | |
| dataAssosiation | 2989 | 65.399 s | 65.399 s | |
| preDefineLandmark | 100 | 34.036 s | 0.084 s | |
| ExtractLandMark | 100 | 6.131 s | 0.327 s | |
| detectLandmark | 100 | 5.046 s | 3.626 s | |
| extractData | 1 | 1.944 s | 0.019 s | |
| imuIntegration | 1 | 1.925 s | 1.925 s | |
| mean | 11799 | 1.878 s | 1.878 s | |
| calculateCentre | 2899 | 1.420 s | 0.355 s | |
| predictionStep | 100 | 0.140 s | 0.093 s | |
| runRobot | 200 | 0.047 s | 0.047 s | |

**Figure 7-22.** Computation time of the SLAM operation without pre-filtering landmarks

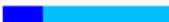

| Function Name | Calls | Total Time | Self Time* | Total Time Plot (dark band = self time) |
|---|---|---|---|---|
| MainProgram | 1 | 11.540 s | 2.795 s | |
| ExtractLandMark | 100 | 4.096 s | 0.143 s | |
| detectLandmark | 100 | 3.567 s | 2.476 s | |
| dataAssosiation | 2042 | 2.467 s | 2.467 s | |
| preDefineLandmark | 100 | 2.249 s | 0.025 s | |
| extractData | 1 | 1.712 s | 0.017 s | |
| imuIntegration | 1 | 1.695 s | 1.695 s | |
| mean | 11799 | 1.234 s | 1.234 s | |
| calculateCentre | 2899 | 1.091 s | 0.403 s | |
| predictionStep | 100 | 0.201 s | 0.126 s | |
| runRobot | 200 | 0.076 s | 0.076 s | |

**Figure 7-23.** Computation time of the SLAM operation with pre-filtering landmarks

Figure 7-24, 7-25, and 7-26 demonstrate the UGV path estimation based on EKF-based SLAM operation, with performing pre-filtering process and 'Landmarks Quality' approach. It can be seen clearly that the EKF-based SLAM operation provide much better path estimation of the UGV in these case, compare to dead reckoning operation. The map builded based on laser scanner observation and robot states estimation, also looks much better and closer to its original form. These results can be used as rapid indicators that the EKF-based SLAM operation implemented in these real data is properly working.

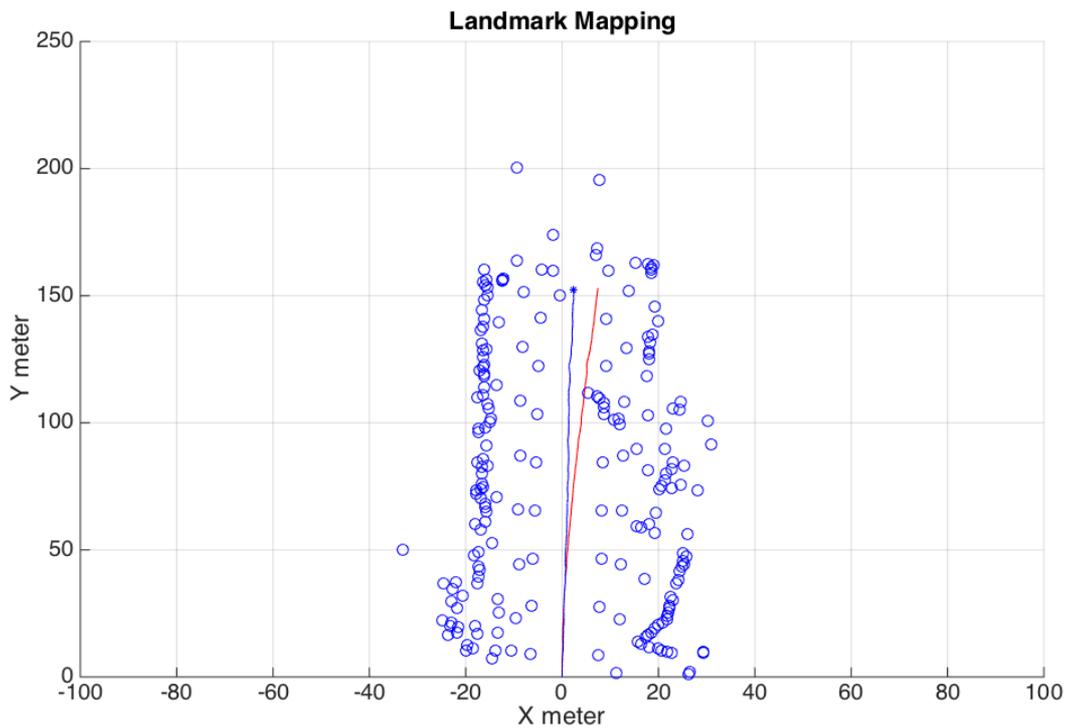

**Figure 7-24**. SLAM operation scenario 1

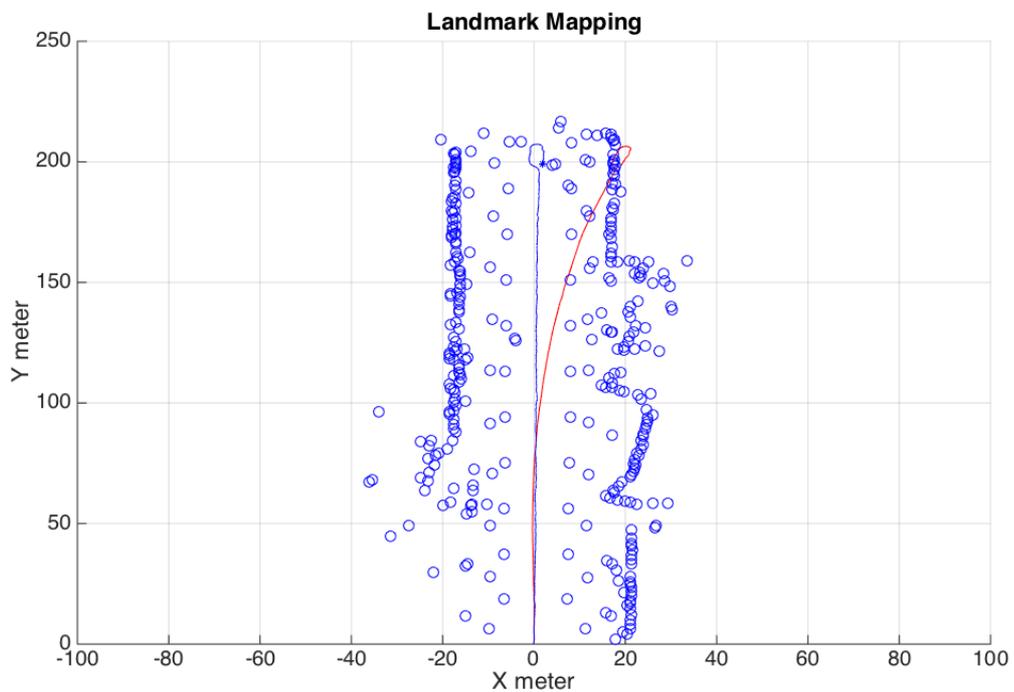

**Figure 7-25**. SLAM operation scenario 2

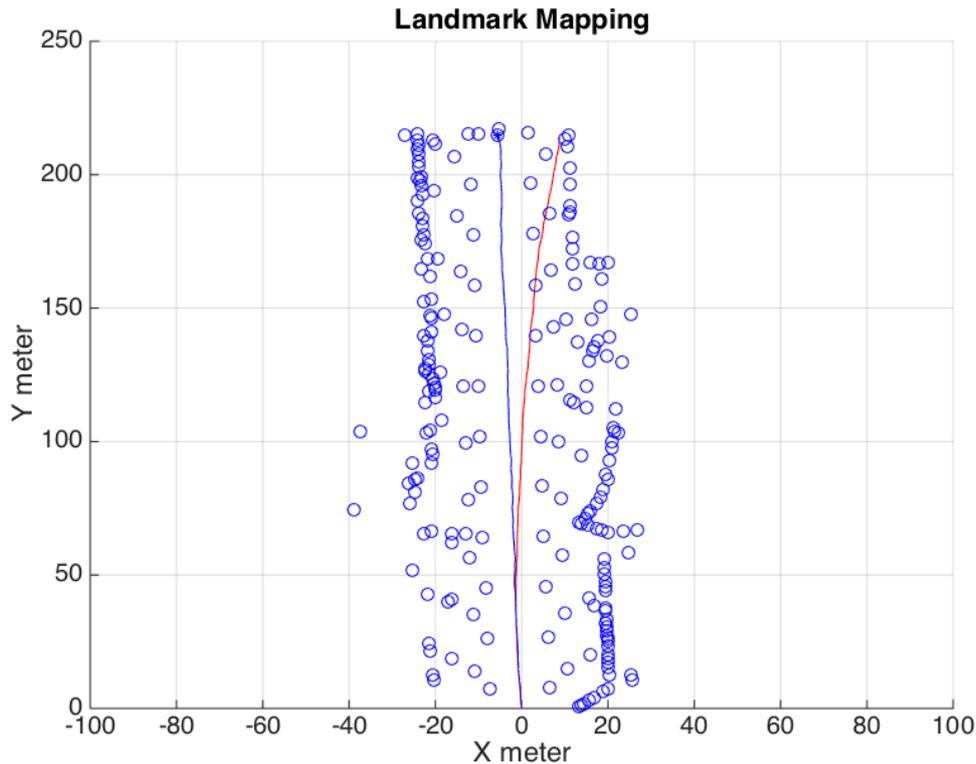

**Figure 7-26**. SLAM operation scenario 3

### 7.5.2. Comparison SLAM operation result and GPS localisation

Another way to evaluate the SLAM operation in this practical process is by comparing the estimated path produced in SLAM operation with the estimated path produced by the GPS receiver. The UGV, used for gathering data for this practical implementation, is equipped by GPS receiver, which record the robot position during the robot travelling. Figure 7-27, 7-28 and 7-29 shows the path estimation based on the GPS receiver, compared to the path estimation generated by SLAM operation. According to the figures it can be seen clearly the problem in the robot localisation based on only GPS system. The estimated position obtained from GPS is jumping frequently and also lack of accuracy. In contrast, from these figure, it clearly can be seen that the SLAM operation produces smoother path and higher accuracy in the estimation robot position. These also can validate that the SLAM operation in this practical implementation is working properly.

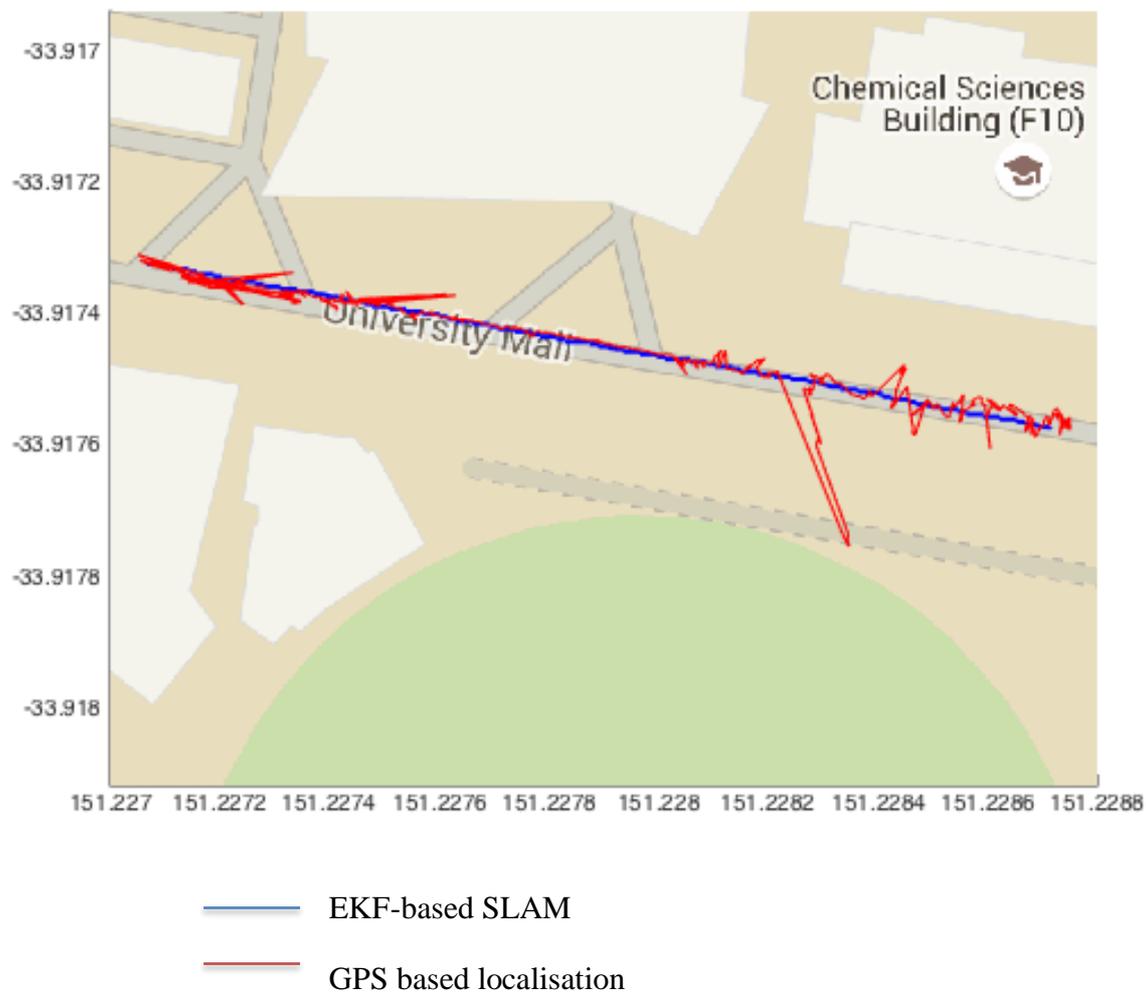

———— EKF-based SLAM

———— GPS based localisation

**Figure 7-27**. GPS localisation compared to SLAM scenario 1

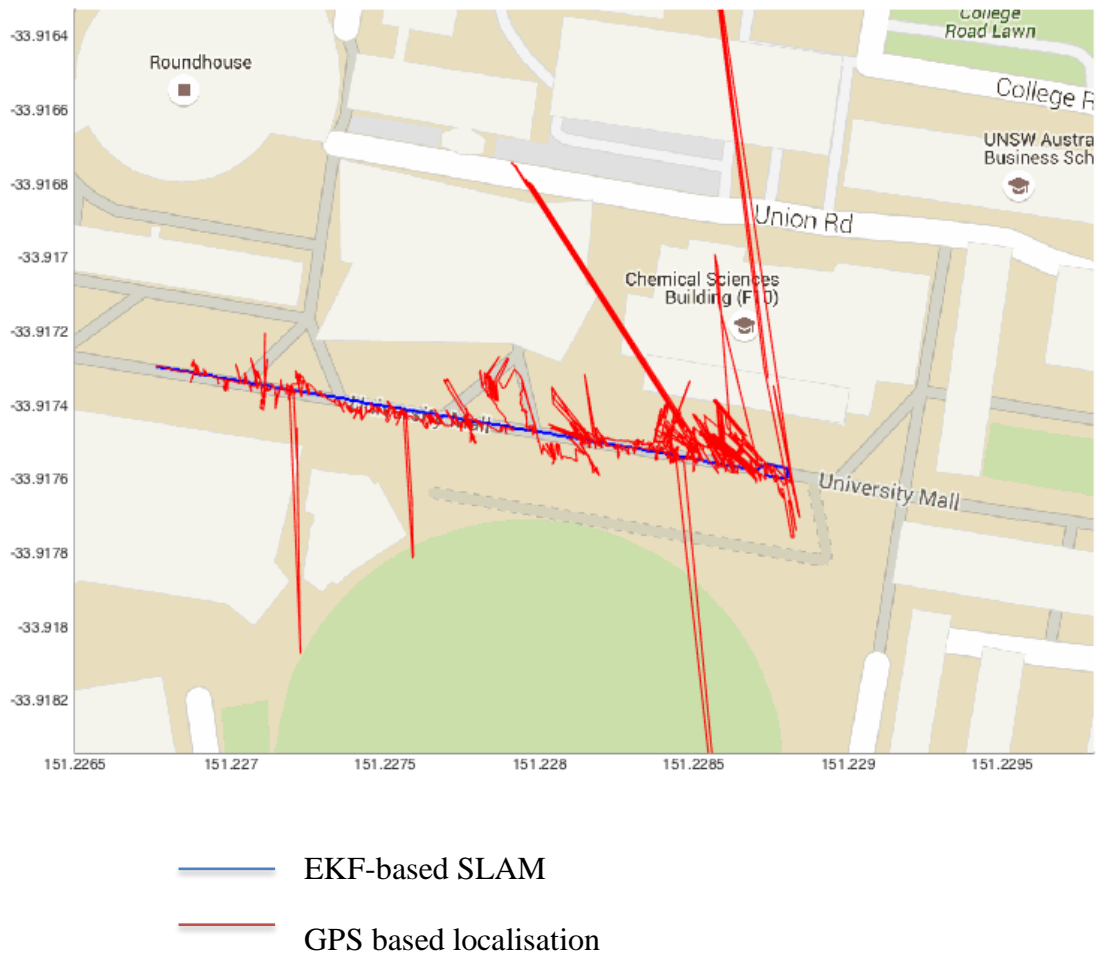

——— EKF-based SLAM

——— GPS based localisation

**Figure 7-28**. GPS localisation compared to SLAM scenario 2

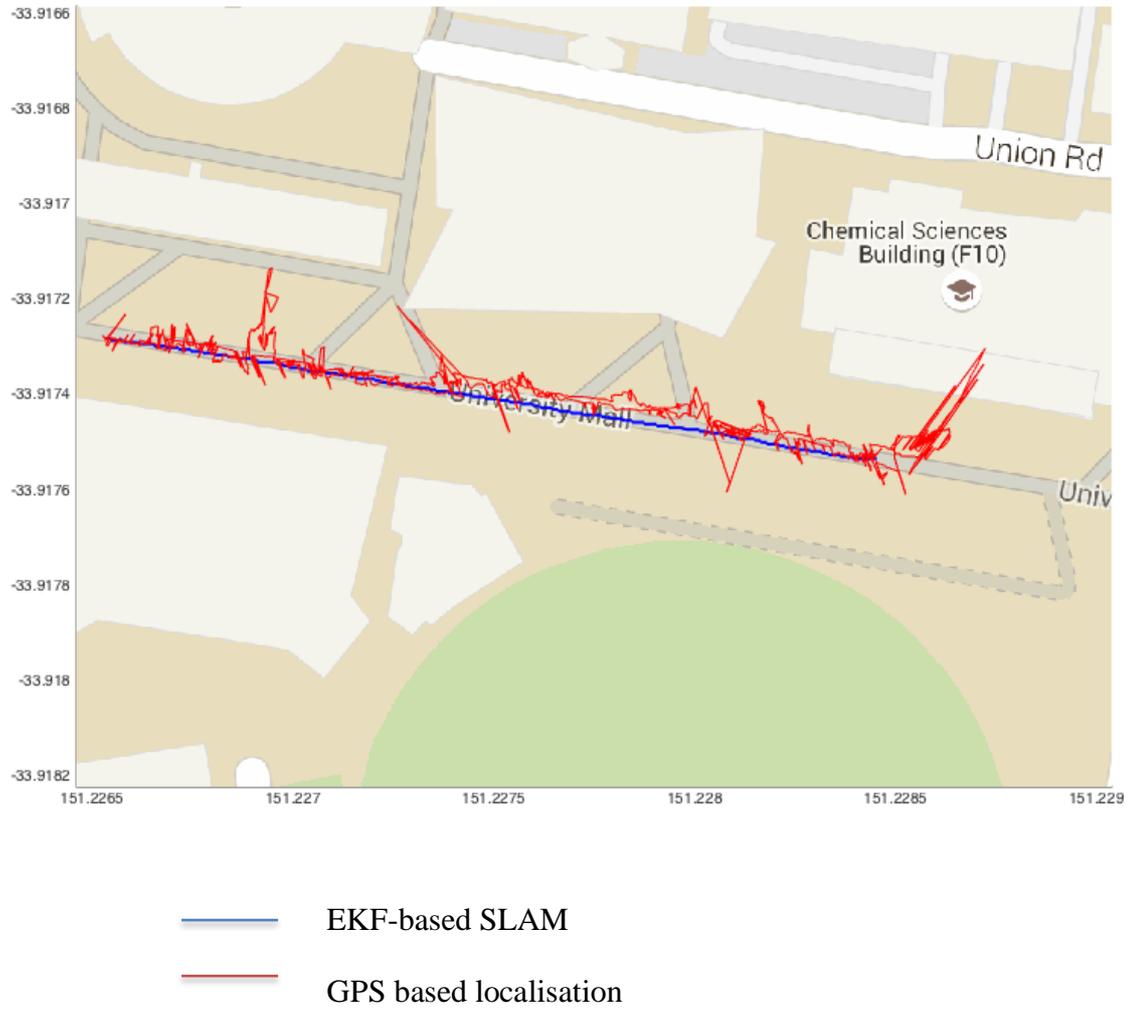

    ———     EKF-based SLAM

    ———     GPS based localisation

**Figure 7-29**. GPS localisation compared to SLAM scenario 3

# Chapter

# 8. Conclusion and Future Recommendations

## 8.1. Conclusion

### 8.1.1. Review

In this project, the implementation of EKF-based 2D SLAM was performed in both simulation process and in practical application using a set of data from the measurement in the real environment. According to the present study, the main conclusion can be stated as follows:

1. EKF-based SLAM implemented in the simulation process demonstrated adequate performance in estimating robot's states and landmark states based on simulation setting.

2. In practical implementation, pre-filtering of landmark extracted from the LMS data was essential to produce good state estimation process based on SLAM operation. Based on the experiment results, it was demonstrated that lack of filtering detected landmark resulted poor quality of state estimation process in SLAM operation, even worse than the state estimations based on dead reckoning technique.

3. Dealing with detected dynamic objects was also essential to produce good result in SLAM operation. Based on experiment in this project, when the dynamic objects were registered as landmarks and were used in SLAM operation, this also would reduce the quality of state estimation based on SLAM operation. In this project, the landmark quality approach was introduced to deal with dynamic objects detected during the SLAM operations. Based on the experiment, after removing poor landmark from database, the state estimation process produced the better results compared to the estimation process without removing these poor landmarks.

4. In this project, The robot's state estimation and map building resulted by SLAM operation was compared to the state estimation and map building resulted by dead reckoning technique and GPS receiver. Based on the experiment result, it showed that the state estimation and map building based on SLAM operation in this project outperformed the dead reckoning technique as well as the GPS based localisation. Using the dead reckoning technique, the estimated robot's state and map building were drifted during the operation. Estimated location resulted by GPS receiver was also lack of accuracy

5. Overall, the practical implementation of 2D EKF-based SLAM operation demonstrated a successful operation accomplishing the goal of this project.

### 8.1.2. Contribution

Generally, this project significantly contributes in to give a comprehensive understanding about the basic implementation of 2D EKF-based SLAM, focusing in practical implementation in the real environment. More specifically, this project is dealing with real environment data, which consists of both static and dynamic objects, at the same time. This project comes out with several practical approaches including pre-filtering extracted landmarks and upgrading and degrading landmark quality to improve the performance of SLAM operation.

## 8.2. Future Recommendations

This present work has contributed to the understanding of practical implementation of EKF-based SLAM in real environment. Based on the obtained findings, we recommend the following future investigation:

- ✓ In this project, all method is implemented and validating only in area around main road of UNSW campus. Therefore, testing all method in the different area is recommended to validate these methods for general cases of application.
- ✓ Future work in optimising the SLAM process to improve computational time and computational cost is also necessary to be explored. Avoiding unnecessary calculation such as multiplying by zero or one can improve efficiency in computational process [**35**] [**30**].
- ✓ Future work on combining GPS data into SLAM operation is also recommended to investigate reliable and affordable SLAM implementation, so that this can be implemented for real mass production application.
- ✓ The measurement of various data from the IMU sensors such as accelerometer and magnetometer is also recommended to investigate the different approach of SLAM operation, and comparing it with the state of the art.

# Appendix 1

## Data Association

```
function [Assoc] = dataAssosiation(A,B)

SizeA=size(A);
SizeB=size(B);
m = SizeA(2);
n = SizeB(2);
JJ=zeros(1,m);

D=zeros(m,n);
for i=1:m,
    for j=1:n,
        D(i,j)= sqrt(((A(1,i)-B(1,j))^2)+((A(2,i)-B(2,j))^2));
    end;
end;

for i=1:m,
    AA=D(i,:);
    AA=(AA)';
    MinDi=min(AA);

    ii=find(AA==MinDi);
    iii=size(ii);
    LL=iii(1);

    for k=1:LL,
        BB=D(:,ii(k));
        MinDk=min(BB);
        II=find(BB==MinDk);
        if II==i,
            if MinDk<1,
                JJ(i)=ii(k);
            else
                JJ(i)=0;
            end;
        else
            JJ(i)=0;
        end;
    end;
end;
return;
```

## Landmarks Extraction

```matlab
function Li = detectLandmark(ranges)
OOI = struct('Angle',{},'Range',{},'Size',{},'Centre',{});
Li = struct('Angle',{},'Range',{},'Size',{},'Centre',{});
n=1;
m=1;
k=0;
Angle=[0]; Range=[ranges(1)];

for j=2:361,
    delta = abs(ranges(j)-ranges(j-1));
    if delta<0.25,
        Angle=[Angle;(j-1)*0.5]; Range=[Range;ranges(j)];
        k=0;
    else
        OOI(n).Angle = Angle; OOI(n).Range = Range;
        OOI(n).Size = length(Angle);
        if OOI(n).Size <8,
            if OOI(n).Size >3,
                Li(m) = OOI(n);
                Li(m).Centre = 
calculateCentre(Li(m).Angle,Li(m).Range,Li(m).Size);
                m=m+1;
            end
        end
        Angle=[(j-1)*0.5]; Range=[ranges(j)];
        n=n+1;
    end
end
OOI(n).Angle = Angle; OOI(n).Range = Range;

if OOI(n).Size <8,
    if OOI(n).Size >3,
        Li(m) = OOI(n);
        Li(m).Centre = 
calculateCentre(Li(m).Angle,Li(m).Range,Li(m).Size);
    end
end

return
```

## Landmark Quality Approach

```
function [preLandmarks,oldLandmarks,goodLandmarks]=
preDefineLandmark(extractedLandmarks,regLandmarks,preLandmarks)
% Prefiltering landmarks
    Xl = extractedLandmarks(:,1);
    Yl = extractedLandmarks(:,2);
    detectedL = [Xl,Yl];
    if ~isempty(regLandmarks)
        % Identifying known landmarks
        [Assoc] = dataAssosiation(detectedL',regLandmarks');
        II = find(Assoc~=0);
        oldLandmarks = extractedLandmarks(II,:);
        oldLandmarks(:,5:6) = regLandmarks(Assoc(II),5:6);
        extractedLandmarks(II,:)=[];
        detectedL(II,:)=[];
        % Processing unknown landmarks
        if ~isempty(preLandmarks)
            [Assoc] =
dataAssosiation(preLandmarks(:,3:4)',detectedL');
            II = find(Assoc==0);
            III = find(Assoc~=0);
            preLandmarks(III,2)=preLandmarks(III,2)+1;
            preLandmarks(II,2)=preLandmarks(II,2)-3;
            [Assoc] =
dataAssosiation(detectedL',preLandmarks(:,3:4)');
            II = find(Assoc==0);
            qlt = ones(length(II),1);
            reg = zeros(length(II),1);
            preLandmarks=[preLandmarks ; [reg qlt
extractedLandmarks(II,:)]];
        else
             qlt = ones(length(detectedL(:,1)),1);
             reg = zeros(length(detectedL(:,1)),1);
             preLandmarks=[reg qlt extractedLandmarks];
        end
        II= preLandmarks(:,2)<-10;
        preLandmarks(II,:)=[];
        III=preLandmarks(:,2)>15;
        goodLandmarks=[preLandmarks(III,3:end)];
    else
        oldLandmarks = [];
        % Processing unknown landmarks
        if ~isempty(preLandmarks)
            [Assoc] =
dataAssosiation(preLandmarks(:,3:4)',detectedL');
            II = find(Assoc==0);
            III = find(Assoc~=0);
            preLandmarks(III,2)=preLandmarks(III,2)+1;
            preLandmarks(II,2)=preLandmarks(II,2)-3;
            [Assoc] =
dataAssosiation(detectedL',preLandmarks(:,3:4)');
            II = find(Assoc==0);
            qlt = ones(length(II),1);
            reg = zeros(length(II),1);
            preLandmarks=[preLandmarks ; [reg qlt
extractedLandmarks(II,:)]];
```

```
        else
            preLandmarks=[];
            qlt = ones(length(detectedL(:,1)),1);
            reg = zeros(length(detectedL(:,1)),1);
            preLandmarks=[reg qlt extractedLandmarks];
        end
        II= preLandmarks(:,2)<-10;
        preLandmarks(II,:)=[];
        III=preLandmarks(:,2)>15;
        goodLandmarks=preLandmarks(III,3:end);
    end

end
```